\documentclass[preprint,12pt]{elsarticle}

\usepackage{amssymb}
\usepackage{amsmath}

\newtheorem{rem}{Remark}
\newtheorem{thm}{Theorem}

\newtheorem{defn}{Definition}
\newtheorem{lem}{Lemma}
\usepackage{color}
\newtheorem{ex}{Example}

 \usepackage{multirow}
\usepackage{mathrsfs} 

\newtheorem{assum}{Assumption}
\newtheorem{prf}{Proof}
\usepackage[caption=false,font=normalsize,labelfont=sf,textfont=sf]{subfig}

\journal{Engineering Applications of Artificial Intelligence}

\begin{document}

\begin{frontmatter}

\title{PINN-Obs: Physics-Informed Neural Network-Based Observer for Nonlinear Dynamical Systems}

\author[uir,lorraine]{Ayoub Farkane\corref{cor1}}
\author[lorraine,inria]{Mohamed Boutayeb}
\author[uir]{Mustapha Oudani}
\author[uir,leeds]{Mounir Ghogho}

\cortext[cor1]{Corresponding author: \texttt{ayoub.farkane@uir.ac.ma}}

\address[uir]{TICLab, International University of Rabat, Rabat 11100, Morocco}
\address[lorraine]{CNRS–CRAN–7039, University of Lorraine, France}
\address[leeds]{Faculty of Engineering, University of Leeds, LS2 9JT, United Kingdom}
\address[inria]{INRIA Nancy—LARSEN, France}

\begin{abstract}
State estimation for nonlinear dynamical systems is a critical challenge in control and engineering applications, particularly when only partial and noisy measurements are available. This paper introduces a novel Adaptive Physics-Informed Neural Network-based Observer (PINN-Obs) for accurate state estimation in nonlinear systems. Unlike traditional model-based observers, which require explicit system transformations or linearization, the proposed framework directly integrates system dynamics and sensor data into a physics-informed learning process. The observer adaptively learns an optimal gain matrix, ensuring convergence of the estimated states to the true system states. A rigorous theoretical analysis establishes formal convergence guarantees, demonstrating that the proposed approach achieves uniform error minimization under mild observability conditions. The effectiveness of PINN-Obs is validated through extensive numerical simulations on diverse nonlinear systems, including an induction motor model, a satellite motion system, and benchmark academic examples. Comparative experimental studies against existing observer designs highlight its superior accuracy, robustness, and adaptability.
\end{abstract}


\end{frontmatter}
\section{Introduction}
 In modern control theory, dynamic systems are fundamental for modeling and analyzing diverse physical, biological, and technological processes.  A critical challenge within these systems is state estimation: the reconstruction of internal (often unmeasurable) states from available output measurements. This concept, known as observability, quantifies the extent to which a system's outputs reveal its internal dynamics. Systems with high observability enable accurate state reconstruction, whereas those with limited observability present significant challenges.

Conventional state estimation methods, including Luenberger observers, Kalman filters, unknown input observers (UIOs), and sliding mode observers (SMOs)  have been extensively employed to address these challenges.  These approaches typically use linear matrix inequalities (LMIs) to design observer gains, transforming stability and convergence conditions into convex optimization problems based on Lyapunov theory. Although effective for linear or mildly nonlinear systems, these approaches become increasingly computationally demanding and less practical for highly nonlinear dynamics, which limits their scalability and overall performance. 

In recent years, deep learning methods have emerged as a promising alternative for handling nonlinear dynamic
systems. For instance, physics-informed neural networks (PINNs) have demonstrated significant potential by embedding the underlying differential equations directly into the neural network training process. 
This method has been applied to designing state observers, particularly for 
Luenberger and Kazantzis–Kravaris/Luenberger (KKL) observers in nonlinear systems. The authors of \cite{niazi2023learning} proposed a supervised PINN-based approach  to learn the nonlinear transformations required for KKL observers, ensuring robust state estimation by enforcing PDE constraints during training. Similarly, PINN-based observer design is extended to discrete-time nonlinear systems in \cite{alvarez2024nonlinear}, formulating the transformation problem via functional equations embedded into the learning process. Beyond KKL observers, \cite{marani2024unsupervised} proposed an unsupervised PINN observer design using contraction theory, allowing the network to learn observer gains purely from physics-based loss terms. Additionally, hybrid approaches have emerged, such as \cite{de2024hybrid}, which combined PINNs with an Unscented Kalman Filter (UKF) for enhanced state estimation, leveraging both physics-informed learning and Bayesian filtering for improved robustness.

In this paper, we propose a novel PINN-based observer design that directly learns the optimal observer gain while ensuring convergence of the estimated state to the true state. Unlike existing PINN-based observer designs \cite{niazi2023learning, alvarez2024nonlinear, ramos2020numerical, de2024hybrid}, which focus on learning nonlinear transformations or inverse mappings, our approach operates directly on the original nonlinear system, eliminating the need for coordinate transformations. By formulating the observer problem as a physics-constrained learning task, our method enforces the neural network to simultaneously learn both the gain matrix and the state estimates that satisfy the observer equation. This leads to an adaptive observer framework, where the optimal gain is determined automatically, ensuring accurate state estimation.  

The key contributions of this work include:
$(i)$ Development of an adaptive observer framework that integrates PINNs with physics-based regularization, guaranteeing that the estimation error converges to zero.  $(ii)$ A model-free approach that does not require system transformation or linearization. This allows for state estimation based solely on the measured system output, even when not all state variables are directly observable. $(iii)$ Rigorous theoretical guarantees, providing a formal convergence analysis that ensures uniform convergence of the estimated state under mild smoothness and detectability conditions.  
$(iv)$ Extensive validation through simulations on a variety of nonlinear systems, including a fifth-order two-phase induction motor model, a satellite motion system, and classical benchmark examples.  

By embedding the observer gain matrix and state estimation into the same learning process, our method leverages physics-constrained loss minimization to automatically determine the optimal gain, thereby ensuring accurate state tracking. The minimization of the output error inherently provides information about the true state, enabling the neural network to self-adapt to the system dynamics. Our approach thus bridges the gap between classical model-based observer design and modern data-driven learning, offering a robust and flexible solution for nonlinear state estimation.  

The remainder of the paper is organized as follows. Section 2 reviews related works on observer design and physics-informed deep learning. Section 3 describes the governing system and the corresponding observer model. In Section 4, we present the convergence analysis of the proposed method. Section 5 details the simulation studies and results, and Section 6 concludes the paper with a discussion of future research directions.

\section{Related Works}
Traditional state estimation methods, such as the Kalman filter and its non-linear extensions \cite{boutayeb1997convergence,boutayeb1999strong,song1992extended}, have long served as the backbone of control theory. Although these methods are effective for linear systems, they face significant challenges when dealing with strong non-linearities, uncertainties, and sparse or noisy measurements. To overcome these limitations, researchers have increasingly explored data-driven approaches that embed physical knowledge into the learning process. One prominent advancement in this direction is the use of Physics-Informed Neural Networks (PINNs).

PINNs were initially introduced to solve forward and inverse problems governed by partial differential equations (PDEs) by incorporating physical laws directly into the training loss \cite{raissi2019physics}. This integration enables the network to achieve high accuracy even with limited data, as the underlying physics guides the learning process. Building on this foundation, recent research has extended the application of PINNs to state estimation and observer design.

For instance, \cite{arnold2021state} proposed a state-space modeling framework based on PINNs that bypasses the need for computationally expensive numerical PDE solvers, thereby enabling real-time applications. Similarly, \cite{delavari2023adaptive} developed an adaptive reinforcement learning-based observer and controller for wind turbines using interval type-II fuzzy systems, showcasing how data-driven observers can effectively handle uncertainties and disturbances in complex engineering systems.

In another study,  \cite{de2024hybrid} introduced a hybrid state estimation framework that combines PINNs with an adaptive Unscented Kalman Filter (UKF). Their approach refines the PINN architecture through hybrid loss functions and Monte Carlo Dropout, leading to significant improvements in tracking both position and velocity. In the aerospace domain,  \cite{varey2024physics} applied PINNs to satellite state estimation, effectively capturing anomalous accelerations, such as those induced by low-thrust propulsion to achieve superior propagation accuracy compared to conventional methods.

Furthermore,  \cite{alvarez2024nonlinear} developed nonlinear discrete-time observers using PINNs, which learn a nonlinear state transformation map within an observer linearization framework, ensuring robust estimation in highly nonlinear settings. In the energy sector,  \cite{falas2023physics} demonstrated that embedding physics-based constraints into PINNs accelerates power system state estimation, resulting in faster convergence and enhanced accuracy over traditional iterative methods.

On the robotics front, \cite{liu2024enhanced} integrated multimodal proprioceptive data with PINNs and an adaptive UKF to achieve real-time state estimation even in the presence of sensor noise and uncertainties. Complementing these studies,  \cite{stiasny2023physics} examined the efficiency of PINNs for time-domain simulations, showing that these methods can be orders of magnitude faster than conventional solvers while maintaining adequate accuracy.

Additionally, the emergence of neural ordinary differential equations (neural ODEs) \cite{chen2018neural} offers a continuous-time framework that naturally extends residual neural networks to dynamic system modeling. Recent work by \cite{uccak2024adaptive} demonstrates the use of support vector regression (SVR) in backstepping controllers for nonlinear systems, while \cite{shanbhag2025machine} propose a machine learning-based observer constrained to Lie groups, addressing state estimation challenges in robotics and aerospace without relying on system linearization.

Overall, while substantial progress has been made in both classical and neural observer designs, a gap remains in developing methods that can robustly handle highly nonlinear dynamics solely from output measurements without resorting to transformations or linearizations. Our work builds upon these advancements by proposing a PINN-based nonlinear observer that integrates sensor data and system dynamics with a systematic parameter optimization strategy, thereby offering enhanced performance and adaptability for a wide range of control applications.
\section{Governing System}
This section defines the system dynamics and presents the corresponding nonlinear state observer design. Our goal is to establish a framework in which, based on a finite number of measurements, the full state of a nonlinear dynamic system can be accurately estimated. In doing so, we also set the stage for introducing our adaptive Physics-Informed Neural Network (PINN) methodology for observer gain optimization.

\subsection{Dynamic System}
Consider the nonlinear dynamic system:
\begin{equation}
\left\{
\begin{array}{l}
\dot{x}(t) = f(x(t), t) + B\,u(t), \\
y(t) = h(x(t)), \\
x(0) = x_{0},
\end{array}
\right.
\label{1}
\end{equation}
where:
\begin{itemize}
    \item \( x: [0, \infty) \rightarrow \mathscr{D} \subset \mathbb{R}^{n_x} \) represents the state vector of the system, with \( \mathscr{D} \) being a subset of \( \mathbb{R}^{n_x} \). The derivative \( \dot{x}(t) \) denotes the rate of change of the state with respect to time.
    \item \( f: \mathscr{D} \times [0, \infty) \rightarrow \mathbb{R}^{n_x} \) is a nonlinear function that describes the system dynamics.
    \item \( B \) is the input matrix.
    \item \( u(t) \) is the control input.
    \item \( y: \mathscr{D} \rightarrow \mathbb{R}^{m} \) denotes the measured output, with \( 1 \leq m \leq n \).
    \item \( x(0) = x_{0} \in \mathscr{D} \) defines the initial state.
\end{itemize}
In our context, although the output function \( h \) may be nonlinear, we consider it as a linear mapping \( h(x)=Cx \), where \( C \) is the output matrix.

To ensure that system (\ref{1}) is well-posed, we assume that the function \( f(\cdot, t) \) is locally Lipschitz continuous:
\begin{assum}
   For every \( t \in [0, \infty) \), the function \( f(\cdot, t): \mathscr{D} \rightarrow \mathscr{D} \) is locally Lipschitz continuous.
   \label{as1}
\end{assum}
Under Assumption~\ref{as1}, the Cauchy-Lipschitz (Picard-Lindelöf) theorem guarantees that system (\ref{1}) has a unique solution for every initial state \( x(0)=x_0 \).

An important question we address is whether the full state of the system can be determined after a finite number of observations. To formalize this, we introduce the concepts of strong and weak detectability:
\begin{defn}\cite{hautus1983strong}
The system (\ref{1}) is \emph{strongly detectable} if the full state can be determined from a finite number of observations for all trajectories.
\end{defn}
\begin{defn}\cite{shu2007detectability}
The system (\ref{1}) is \emph{weakly detectable} if the full state can be determined from a finite number of observations for some trajectories.
\end{defn}
Detectability has been further characterized using tools such as the Riemann metric \cite{sanfelice2011convergence,sanfelice2015convergence,sanfelice2023convergence} and information geometry \cite{amari2000methods}, which assess the distinguishability of states on an information manifold.

\subsection{Nonlinear State Observer}
A state observer (or estimator) is a model that reconstructs the internal states of a dynamic system from its inputs and outputs. For system (\ref{1}), a corresponding nonlinear state observer can be formulated as
\begin{equation}
\left\{
\begin{array}{l}
\dot{\hat{x}}(t)=f(\hat{x}(t), t) + L\,C\left(x(t)- \hat{x}(t)\right) + B\,u(t), \\
\hat{x}(0)=\hat{x}_{0},
\end{array}
\right.
\label{3}
\end{equation}
where \( \hat{x}(t) \) is the estimated state vector and \( L \) is the observer gain matrix. In time-varying observers, \( L \) may be a function of time:
\begin{equation}
\left\{
\begin{array}{l}
\dot{\hat{x}}(t)=f(\hat{x}(t), t) + L(t)\,C\left(x(t)- \hat{x}(t)\right) + B\,u(t), \\
\hat{x}(0)=\hat{x}_{0}.
\end{array}
\right.
\label{4}
\end{equation}
The objective is to design a gain matrix \( L(t) \) such that the estimation error \( e(t) = x(t) - \hat{x}(t) \) converges to zero as \( t \to \infty \) regardless of the initial conditions \( x_0 \) and \( \hat{x}_0 \). In other words, we wish to achieve asymptotic stability of the set
\begin{equation}
\mathcal{S} := \{ (x, \hat{x}) \in \mathscr{D} \times \mathscr{D} : x = \hat{x} \}.
\end{equation}

Traditional techniques such as those based on the Lyapunov equation, linear matrix inequalities (LMIs) \cite{alessandri2004design}, or Kalman filter formulations provide systematic ways to compute \( L \) for linear or mildly nonlinear systems. However, for highly nonlinear dynamics or systems with weak observability, these methods may not be applicable.

\subsection{Proposed Adaptive PINN-Based Model}
\begin{figure*}[!t]
     \centering
     \includegraphics[width=\textwidth]{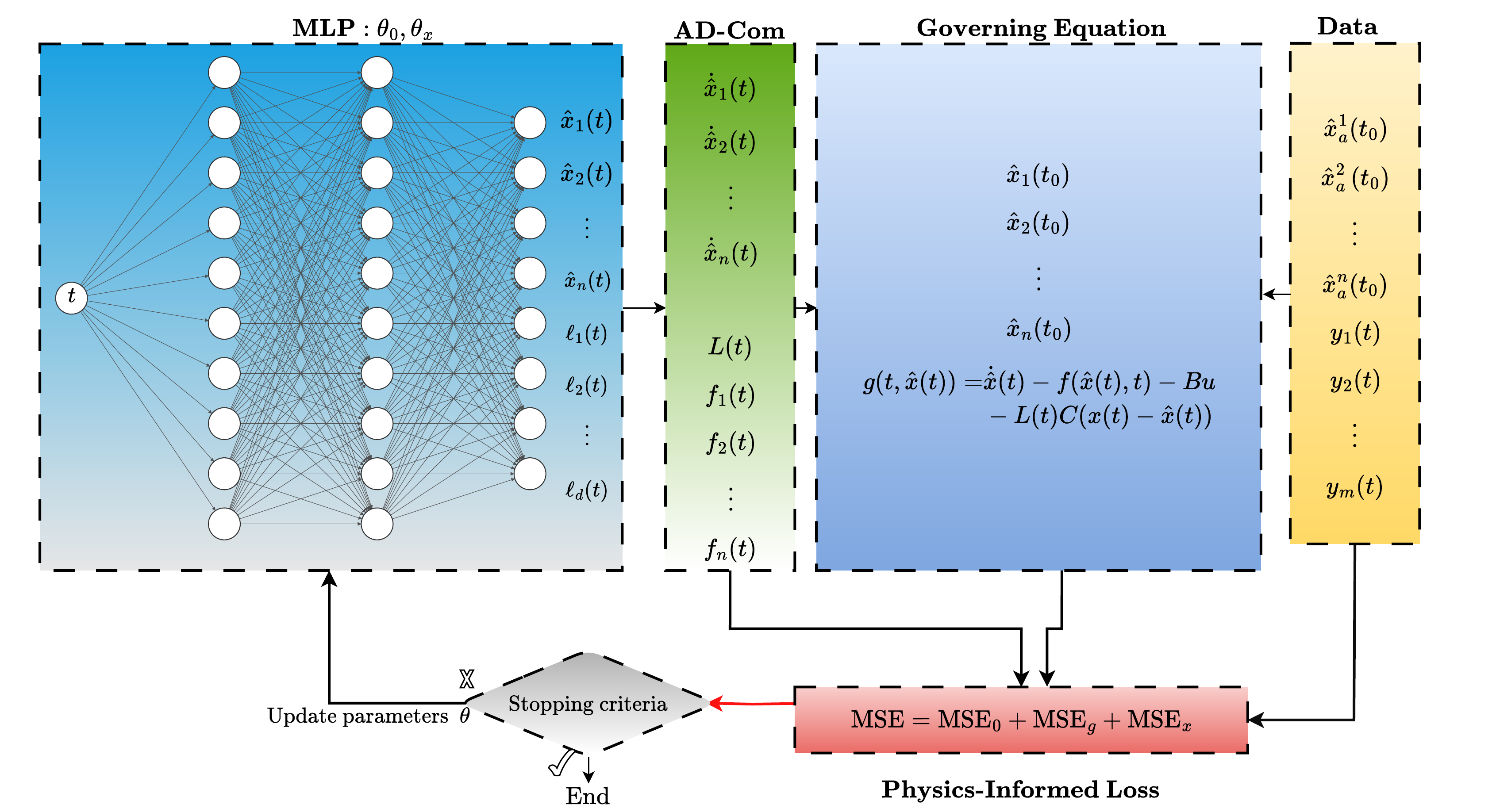}
     \caption{Adaptive PINNS for nonlinear state observer design. }
     \label{f1}
 \end{figure*}
To overcome these challenges, we propose an adaptive PINN-based approach for determining the optimal observer gain \( L(t) \). By leveraging the high accuracy of PINNs in approximating complex nonlinear functions, our method embeds the observer design into a neural network framework. A Multi-Layer Perceptron (MLP) receives time instances \( t=(t_0,t_1,\dots,t_N) \) as inputs and outputs both the candidate state estimate \( \hat{x}(t) \) and a set of gain coefficients \( [l_1, l_2, \dots, l_d]^T \), which are reshaped into the gain matrix
\[
L(t)=\begin{pmatrix}
l_1 & l_2 & \dots & l_m\\
l_{m+1} & l_{m+2} & \dots & l_{2m}\\
\vdots & \vdots & \ddots & \vdots\\
l_{n} & l_{n+1} & \dots & l_{d}
\end{pmatrix}.
\]
After generating the candidate solution, Automatic Differentiation is employed to compute the time derivative \( \dot{\hat{x}}(t) \). The governing equation is then formed as:
\begin{equation}
g(t,\hat{x},\theta)=\dot{\hat{x}}(t)-f(\hat{x}(t),t)-B\,u(t)-L(t)C\left(x(t)-\hat{x}(t)\right),
\label{e55}
\end{equation}
which quantifies the mismatch between the observer dynamics and the actual system behavior. Data from real sensors is used to optimize the following composite mean squared error (MSE) loss function:
\begin{equation}
\text{MSE}=\text{MSE}_0 + \text{MSE}_g + \text{MSE}_y,
\end{equation}
where
\begin{align}
    \text{MSE}_0 &= \|\hat{x}(t_0)-\hat{x}_a(t_0)\|, \label{7}\\[1mm]
    \text{MSE}_g &= \frac{1}{N}\sum_{i=1}^{N}\|g(t_g^i,\hat{x}_g^i)\|^2, \label{8}\\[1mm]
    \text{MSE}_y &= \frac{1}{N}\sum_{i=1}^{N}\|\hat{y}(t^i)-y(t^i)\|^2. \label{9}
\end{align}
Here, \( \hat{x}(t_0) \) is the candidate initial state, \( \hat{x}_a(t_0) \) is the actual initial estimate, and \( \hat{y}(t) \) is the estimated output computed as \( \hat{y}(t) = C\,\hat{x}(t) \).

\subsection{Neural Network Training and Testing}
The training data typically consists of tuples \( \{ t^i, y(t^i), \hat{x}_a(t_0) \} \) obtained either from real-world measurements or numerical ODE solvers. The network is trained by minimizing the loss defined above, which ensures that:
\begin{itemize}
    \item \(\text{MSE}_0\) enforces the correct initialization of the observer,
    \item \(\text{MSE}_g\) drives the adaptive gain \( L(t) \) to enforce convergence of the observer dynamics,
    \item \(\text{MSE}_y\) aligns the estimated output with the measured system output.
\end{itemize}

During testing, new time instances \( t \) are used along with sensor data \( y(t) \) and the previously trained observer gains \( L(t) \) to numerically solve the observer ODE system.  The accuracy of the observer is assessed by comparing the true state \( x_i(t) \) with the estimated state \( \hat{x}_i(t) \) for each variable, where the instantaneous error is defined as

\[
\text{Error}_i = \big| x_i(t) - \hat{x}_i(t) \big|.
\]

To quantify the overall performance, we introduce several metrics. The Mean Absolute Error (MAE) is given by

\[
\text{MAE} = \frac{1}{n} \sum_{i=1}^{n} \big| x_i(t) - \hat{x}_i(t) \big|,
\]

which represents the average of the absolute differences between the true and estimated states. To penalize larger deviations more severely, the Mean Squared Error (MSE) is defined as

\[
\text{MSE} = \frac{1}{n} \sum_{i=1}^{n} \big( x_i(t) - \hat{x}_i(t) \big)^2.
\]

Taking the square root of the MSE yields the Root Mean Squared Error (RMSE):

\[
\text{RMSE} = \sqrt{\text{MSE}},
\]

which brings the error measure back to the original units of the state variables and facilitates interpretation. Furthermore, the Symmetric Mean Absolute Percentage Error (SMAPE) is introduced as a relative error metric and is mathematically expressed as

\[
\text{SMAPE} = \frac{100\%}{n} \sum_{i=1}^{n} \frac{\big| x_i(t) - \hat{x}_i(t) \big|}{\frac{\big| x_i(t) \big| + \big| \hat{x}_i(t) \big|}{2}},
\]
which provides an error percentage 
by normalizing the absolute error with the 
average of the magnitudes of the true and estimated states. Collectively, these metrics MAE, MSE, RMSE, and SMAPE offer a comprehensive assessment of the observer's performance, capturing both the absolute and relative accuracies of the state reconstruction. Lower values of these metrics 
indicate a more precise estimation of the system's states.

\begin{rem}
    In some cases, the gain coefficients in \( L(t) \) may be repetitive, which implies that the system may have a constant observer gain.
\end{rem}

\subsection{Discussion}
The adaptive PINN-based approach presented here offers both theoretical and practical contributions. Theoretically, it embeds the observer design into a unified neural network framework that directly approximates the observer gain \( L(t) \) alongside the state estimate \( \hat{x}(t) \). Practically, by incorporating real sensor data and using Automatic Differentiation, the method provides a flexible and computationally efficient solution for nonlinear state estimation even in cases where the nonlinear function \( f \) is not globally Lipschitz or the system is only weakly observable.

\section{Convergence Analysis}
In this section, we provide a theoretical characterization of our proposed adaptive PINN-based observer design by formulating an optimization problem. Our goal is to show that, under appropriate conditions and with sufficient training data, the estimated state vector \( \hat{x}(t) \) converges to the actual state \( x(t) \) as \( t \to \infty \).

\subsection{Formulation of the Optimization Problem}
Recall the state observer defined in (\ref{4}), which is designed so that the estimation error \( e(t)=x(t)-\hat{x}(t) \) converges to zero over time. The adaptive PINN framework achieves this by simultaneously approximating the observer gain \( L(t) \) and the state estimate \( \hat{x}(t) \) through a neural network. To this end, we define the composite loss function as
\begin{equation}
\text{MSE} = \text{MSE}_0 + \text{MSE}_g + \text{MSE}_y,
\label{10}
\end{equation}
where:
\begin{itemize}
    \item \(\text{MSE}_0\) measures the error in matching the initial condition,
    \item \(\text{MSE}_g\) quantifies the residual error in the governing observer dynamics (i.e., how well the network output satisfies the differential equation), and
    \item \(\text{MSE}_y\) measures the discrepancy between the estimated output \( \hat{y}(t)=C\hat{x}(t) \) and the measured output \( y(t)=Cx(t) \).
\end{itemize}

The loss \(\text{MSE}\) depends on the neural network parameters (i.e., weights and biases) and, implicitly, on the observer gain \( L(t) \) produced by the network. Given a set of \( N \) training samples, our training dataset is defined as
\[
\mathcal{T}_N = \{(t^i, y(t^i))\}_{i=1}^{N} \cup \{\hat{x}_a(t_0)\},
\]
where \( t^i \) are time instances, \( y(t^i) = Cx(t^i) \) are the measured outputs, and \( \hat{x}_a(t_0) \) is the initial estimated state.

For a given class of neural networks \( \mathcal{NN}_{\ell} \) with \(\ell\) parameters, we seek a minimizer \( h^* \) that satisfies
\begin{align}
\text{MSE}(h, t, \hat{x}_0) &= \text{MSE}_0(h, t, \hat{x}_0) + \text{MSE}_g(h, t, \hat{x}_0) + \text{MSE}_y(h, t, \hat{x}_0),
\label{11}
\end{align}
where \( h \) represents a candidate neural network mapping from the time variable (and initial condition) to the pair \( (\hat{x}(t), L(t)) \). To formalize the optimization, we define a discrete probability measure on the training set:
\[
\mu_N = \frac{1}{N} \sum_{i=1}^{N} \delta_{t^i}.
\]
Then, the expected loss is given by
\begin{equation}
\operatorname{Loss}_{N}(h; t, \hat{x}_0) = \mathbb{E}_{\mu_N}\left[ \operatorname{MSE}(h, t, \hat{x}_0) \right].
\end{equation}
This leads to the following minimization problem:
\begin{equation}
\min_{h \in \mathcal{NN}_{\ell}} \operatorname{Loss}_{N}(h; t, \hat{x}_0).
\label{p17}
\end{equation}

\begin{rem}
    It is important to note that the minimization problem defined in (\ref{p17}) may not have a globally optimal solution due to the non-convexity of the loss function. However, for any given tolerance \( \epsilon > 0 \), there exists a suboptimal solution \( h^\epsilon \in \mathcal{NN}_{\ell} \) such that
    \begin{equation}
    \operatorname{Loss}_{N}(h^\epsilon) \leq \inf_{h \in \mathcal{NN}_{\ell}} \operatorname{Loss}_{N}(h) + \epsilon.
    \end{equation}
    This guarantee of \(\epsilon\) optimality ensures that our adaptive PINN can approximate the observer gain and state estimate with arbitrarily small errors given sufficient model capacity and training data.
\end{rem}
\subsection{Neural Network Configuration}
We consider a feed-forward neural network \( h^k: \Omega \rightarrow \mathbb{R}^{d_{\text{out}}} \) with \( k \) layers, where \( \Omega = [0, T] \). The \( k \)-th layer comprises \( l_{k} \) neurons with weights \( \boldsymbol{W}^k \in \mathbb{R}^{l_k \times l_{k-1}} \) and biases \( \boldsymbol{b}^k \in \mathbb{R}^{l_k} \); we denote the parameters of the network by
\[
\theta_k = \{(\boldsymbol{W}^j, \boldsymbol{b}^j)\}_{j=1}^{k}.
\]
For notational clarity, we set \( l_0=1 \) and \( l_k=d_{\text{out}} \). The network architecture is specified by the vector \( \vec{\ell}=(l_0, l_1, \dots, l_k) \in \mathbb{N}^{k+1} \). The network is defined recursively:
\[
h^{k'}(\mathbf{t}) = \boldsymbol{W}^{k'} \, \sigma\left(h^{k'-1}(\mathbf{t})\right) + \boldsymbol{b}^{k'}, \quad \text{for } 2 \leq k' \leq k,
\]
with \( h^1(\mathbf{t}) = \boldsymbol{W}^1 \mathbf{t} + \boldsymbol{b}^1 \) and input \( \mathbf{t} \in \mathbb{R}_{+}^{l_0} \). The overall network is denoted by
\[
h^k\left(\mathbf{t}; \vec{\ell}, \theta_k\right),
\]
or simply \( h^k(\mathbf{t}; \theta_k) \) when the context is clear.

We define the function class of neural networks as
\[
\mathcal{NN}_{\vec{\ell}} = \left\{h^k\left(\cdot; \vec{\ell}, \theta_k\right): \Omega \rightarrow \mathbb{R}^{d_{\text{out}}} \mid \theta_k = \{(\boldsymbol{W}^j, \boldsymbol{b}^j)\}_{j=1}^k \right\}.
\]
The optimization problem in (\ref{p17}) is thus recast as
\begin{equation}
\min_{\theta_k} \operatorname{MSE}(\theta_k; t, \hat{x}_0).
\label{op14}
\end{equation}
This problem is typically solved using gradient descent methods (e.g., L-BFGS, Adam) following these steps:
\begin{enumerate}
    \item \textbf{Initialization:} Set initial parameters \( \theta \).
    \item \textbf{For each training example} \( (t^i, y(t^i)) \):
    \begin{itemize}
        \item Compute \( h(t^i, \theta) \).
        \item Update \( \theta \leftarrow \theta - \eta \nabla_{\theta} \text{Loss}(\theta, t^i, \hat{x}_0) \), where \( \eta \) is the learning rate.
    \end{itemize}
    \item \textbf{Iteration:} Repeat until convergence.
\end{enumerate}

A critical component in our architecture is the choice of activation function. We employ the hyperbolic tangent (tanh) function defined as
\[
\tanh(x) = \frac{e^x - e^{-x}}{e^x + e^{-x}},
\]
which is bounded in \((-1,1)\), is 1-Lipschitz continuous (since its derivative \(1-\tanh^2(x)\) is bounded by 1), and whose higher-order derivatives can be expressed as polynomials in \( \tanh(x) \).

To estimate the state \( \hat{x}(t) \) and the observer gain \( L(t) \), the neural network output is structured as
\[
h(t) = \begin{bmatrix} h_1(t) \\ h_2(t) \end{bmatrix} \equiv \begin{bmatrix} \hat{x}(t) \\ L(t) \end{bmatrix}.
\]
Moreover, we assume that \( h \) belongs to a Hölder space \( C^{k, \beta}(\bar{\Omega}) \) with exponent \( \beta \) (see the definitions below), which guarantees a certain level of smoothness necessary for convergence.

\begin{defn}
Let \( \Omega \) be a bounded domain and \( \beta \in (0,1] \). For \( f \in C^k(\bar{\Omega}) \), the Hölder seminorm is defined as
\[
[f]_\beta = \sup_{x,y\in\bar{\Omega},\, x\neq y} \frac{\|f(x)-f(y)\|}{|x-y|^\beta},
\]
and the Hölder norm by
\[
\|f\|_{C^{k, \beta}(\bar{\Omega})} = \sum_{|\alpha|\leq k} \|D^\alpha f\|_{C^0(\bar{\Omega})} + \sum_{|\alpha|=k} [D^\alpha f]_\beta.
\]
The space \( C^{k, \beta}(\bar{\Omega}) \) consists of functions for which this norm is finite.
\end{defn}

\begin{assum}\label{as2}
We assume that there exists a constant \( \beta > 0 \) such that the neural network \( h \) belongs to the Hölder space \( C^{k,\beta'}(\bar{\Omega}) \) for all \( \beta' \in (0,\beta] \).
\end{assum}

\begin{lem}\label{l1}
Let \( x \in \mathscr{D}\subset \mathbb{R}^n \) and \( y \in \mathbb{R}^m \). Suppose \( h:\Omega \rightarrow \mathbb{R}^{d_{\text{out}}} \) is a neural network in \( \mathcal{NN}_{\vec{\ell}} \) with \( h \in C^{k, \beta}(\bar{\Omega}) \) for some \( \beta \in (0,1] \). Assume further that \( y(\cdot, t) \) is continuously Lipschitz and \( f(\cdot, t) \) is locally Lipschitz. Then, there exists a constant \( C(h) < \infty \) such that the Hölder norm of the residual function \( g[h](t) \) defined by
\[
g[h](t)= \frac{d}{dt}h_1(t) - f(h_1(t),t) - h_2(t)\big(y(x(t))- y(h_1(t))\big)
\]
satisfies
\[
[g[h]]_{\beta;\Omega}^2 \leq C(h).
\]
\end{lem}

\begin{prf}
The proof is based on the Hölder continuity of \( h \) and the Lipschitz properties of \( f \) and \( y \). By decomposing the difference \( g[h](t_1)-g[h](t_2) \) into terms associated with \( \frac{d}{dt}h_1 \), \( f \), and the product \( h_2(y(x)-y(h_1)) \), and using the uniform boundedness of \( h_2 \) and the Lipschitz continuity of \( f \) and \( y \), one can bound each term. Applying the definition of the Hölder seminorm then leads to the desired inequality.
\end{prf}

\begin{assum}
Assume that the training samples \( \mathcal{T}_N \) are drawn from a probability distribution \( \mu_N \) whose density \( \rho \) is supported on \( \bar{\Omega} \) with \( \inf \rho > 0 \).
\end{assum}

\begin{lem}[Theorem 3.1 in \cite{shin2020convergence}]\label{l2}
Suppose the conditions of Lemma \ref{l1} hold. Then, with probability at least 
\[
1 - \sqrt{N}\left(1 - \frac{1}{\sqrt{N}}\right)^N,
\]
the following inequality holds:
\[
\operatorname{Loss}(h; t, \hat{x}_0) \leq C_N \cdot \operatorname{Loss}_N(h; t, \hat{x}_0) + C' N^{-\beta},
\]
where \( C' \) is a constant depending on \( \beta \), \( f \), and \( x \), and \( C_N \) depends on \( N \).
\end{lem}

This result ensures that minimizing the empirical loss \( \operatorname{Loss}_N \) over the network class \( \mathcal{NN}_{\vec{\ell}}^N \) approximates the expected loss \( \operatorname{Loss}(h; t, \hat{x}_0) \) as \( N \to \infty \). We assume that for each \( N \) there exists a neural network \( H^* \in \mathcal{NN}_{\vec{\ell}}^N \) such that \( \operatorname{Loss}_N(H^*, t, \hat{x}_0) = 0 \).

\begin{thm}
Let \( \Omega \) be a bounded subset of \( \mathbb{R}_+ \), and suppose \( x \in C^{0,\beta} \), \( f(\cdot, t) \), and \( y(\cdot, t) \) satisfy the smoothness assumptions. Assume that the system (\ref{1}) is weakly detectable. Let \( h_N \in \mathcal{NN}_{\vec{\ell}}^N \) be a minimizer of the optimization problem (\ref{op14}) with \( h_N = [h_N^1 \,\, h_N^2]^T \), where \( h_N^1 \) represents the state estimate and \( h_N^2 \) corresponds to the observer gain \( L(t) \). Then:
\begin{enumerate}
    \item For any chosen gain matrix \( L \), the observer system (\ref{4}) admits a unique solution \( \hat{x}^* \) for any initial estimate \( \hat{x}_0 \).
    \item With probability 1 over the training set \( \mathcal{T}_N \), it holds that
    \[
    \lim_{N\to\infty} h_N^1 = \hat{x}^* \quad \text{in } C^0(\Omega).
    \]
\end{enumerate}
\end{thm}
\begin{prf}
We prove the theorem in two main steps.

\textbf{Step 1. Existence and Uniqueness of the Observer Solution:}  
By Assumption (1) and the Cauchy-Lipschitz theorem, system (\ref{1}) possesses a unique solution for any initial condition. Under the weak detectability assumption, there exists an observer gain \( L(t) \) such that the observer system (\ref{4})
\[
\dot{\hat{x}}(t)=f(\hat{x}(t),t)+L(t)C\bigl(x(t)-\hat{x}(t)\bigr)+Bu(t)
\]
has a unique solution \( \hat{x}^*(t) \) for any initial estimated state \( \hat{x}_0 \).

\textbf{Step 2. Convergence of the Neural Network Approximation:}  
By construction, \( h_N \) minimizes the empirical loss
\[
\operatorname{Loss}_N(h; t, \hat{x}_0) = \mathbb{E}_{\mu_N}\left[\text{MSE}(h; t, \hat{x}_0)\right].
\]
Since the loss function \eqref{11} is a sum of nonnegative terms, if \( \operatorname{Loss}_N(h_N; t, \hat{x}_0) \to 0 \) as \( N\to\infty \), then each of its terms converges to zero. In particular, we obtain:
\[
\lim_{N\to\infty} \| h_N^1(t_0) - \hat{x}_0 \| = 0, \quad \lim_{N\to\infty} \| g[h_N](t) \| = 0, \quad \text{and} \quad \lim_{N\to\infty} \| C\,h_N^1(t) - y(t) \| = 0,
\]
uniformly over \( t \in \Omega \).

Next, by Lemma~\ref{l1}, the residual function \( g[h_N] \) has a uniformly bounded Hölder seminorm:
\[
[g[h_N]]_{\beta; \Omega} \leq C(h_N) < \infty.
\]
Since the network functions \( h_N \) lie in the Hölder space \( C^{k,\beta}(\bar{\Omega}) \) (by assumption), the sequence \( \{ h_N^1 \} \) is uniformly bounded and equicontinuous. By the Arzelà-Ascoli theorem, there exists a subsequence (which we denote again by \( h_N^1 \) for simplicity) that converges uniformly in \( C^0(\Omega) \) to a limit function \( g^1 \).

Because the empirical loss converges to zero, it follows that:
\[
\|g[h_N]\|_{C^0(\Omega)}^2 + \|h_N^1(t_0)-\hat{x}_0\|^2 + \|C\,h_N^1(t) - y(t)\|^2 \to 0 \quad \text{as } N\to\infty.
\]
Passing to the limit (using the uniform convergence of \( h_N^1 \) to \( g^1 \) and continuity of \( f \) and \( y \)), we deduce that:
\[
\frac{d}{dt} g^1(t) - f\bigl(g^1(t),t\bigr) - L(t)C\Bigl(x(t)-g^1(t)\Bigr) = 0, \quad g^1(t_0) = \hat{x}_0, \quad \text{and} \quad C\,g^1(t) = y(t).
\]
By the uniqueness of the solution to (\ref{4}), it follows that \( g^1(t) = \hat{x}^*(t) \) for all \( t \in \Omega \). Thus, the entire sequence \( h_N^1 \) converges uniformly to \( \hat{x}^* \) in \( C^0(\Omega) \).

This completes the proof.
\end{prf}

\subsection{Discussion}
The convergence analysis demonstrates that by minimizing the combined loss function (data loss, physics loss, and output matching loss), our adaptive PINN-based observer framework converges to a unique solution \( \hat{x}^* \) that satisfies the system dynamics. The theoretical results hinge on key assumptions such as the local Lipschitz continuity of \( f \), the Hölder continuity of the neural network approximator, and the weak detectability of the system. Importantly, the established lemmas and theorem ensure that, with an increasing number of training samples and under sufficient network complexity, the observer’s state estimate converges in \( C^0(\Omega) \) to the true state of the system.

This analysis confirms that our approach is theoretically sound: it guarantees that the neural network, by approximating both the observer gain and the state, yields an error that can be made arbitrarily small. These results provide a solid foundation for the practical implementation and further numerical validations of the proposed adaptive PINN-based state observer.
\section{Applications}
In this section, we assess the efficacy of the proposed adaptive PINN-based observer in various challenging scenarios involving highly nonlinear and complex system dynamics. In practice, selecting an appropriate neural network architecture requires balancing model complexity, training efficiency, and generalization capability \cite{kuri2014best, oh2021optimal}. Our experiments follow an iterative process of experimentation and refinement to achieve an optimal configuration.
\subsection{Ablation studies}
We analyze the Reverse Duffing Oscillator defined by:
\begin{equation}
\left\{
\begin{array}{l}
\dot{x}_1 = x_2^3,\\[1mm]
\dot{x}_2 = -x_1,\\[1mm]
y = x_1.
\end{array}
\right. \label{reverse}
\end{equation}
Notably, the system's trajectories remain confined within invariant compact sets. However, its weak differential observability significantly complicates state estimation, making it challenging to reconstruct the complete state from the measured output \cite{ramos2020numerical,peralez2021deep,buisson2023towards,niazi2023learning}.
We apply the adaptive PINN-based observer to estimate the state component \(x_2\) based on measurements of \(x_1\). The neural network is trained under the following configuration: the initial state is set to \(x_0 = [2, -1]\) with an initial estimate \(\hat{x}_0 = [1, 1]\), and the control input is \(u = 0\). The simulation runs for a maximum duration of \(T_{\text{max}} = 20\,\text{s}\) with a time step of \(\delta_t = 2 \times 10^{-3}\,\text{s}\). Additionally, the training is configured to run for up to 200,000 iterations with an early stopping criterion set to a patience of 20,000 iterations.
\subsubsection{Neural  Network Depth}
\begin{table}[ht]
\centering
\caption{Performance Evaluation of the Adaptive PINN-Based Observer: Impact of Varying Layers and Neuron Counts. The table lists RMSE, MAE, inference time (I-Time in ms), training time (T-Time in s), convergence iterations, and the best achieved loss value.}
\label{tab:performance_metrics}
\resizebox{\textwidth}{!}{%
\begin{tabular}{ccllclll}
\hline
\textbf{Layers}             & \textbf{Neurons} & \multicolumn{1}{c}{\textbf{RMSE}}   & \multicolumn{1}{c}{\textbf{MAE}}     & \textbf{\begin{tabular}[c]{@{}c@{}}I- Time \\ (ms)\end{tabular}} & \multicolumn{1}{c}{\textbf{\begin{tabular}[c]{@{}c@{}}T-Time\\  (s)\end{tabular}}} & \multicolumn{1}{c}{\textbf{\begin{tabular}[c]{@{}c@{}}Conv. \\ Iter.\end{tabular}}} & \multicolumn{1}{c}{\textbf{\begin{tabular}[c]{@{}c@{}}Best \\ loss\end{tabular}}} \\ \hline
\multirow{4}{*}{15}         & 30               & 0.0629                              & 0.01409                              & 2.4250                                                           & 2445.49                                                                            & 71872                                                                               & $1.21\times10^{-3}$                                                                          \\
                            & 20               & 0.0648                              & 0.02211                              & 9.9559                                                           & \multicolumn{1}{c}{2856.84}                                                        & \multicolumn{1}{c}{87365}                                                           & $1.54\times10^{-3}$                                                      \\
                            & 15               & 0.2422                              & 0.05898                              & \multicolumn{1}{l}{1.5795}                                       & 3509.59                                                                            & 119552                                                                              & $4.29\times10^{-3}$                                                                          \\
                            & 10               & 0.2792                              & 0.09601                              & \multicolumn{1}{l}{1.3609}                                       & 5092.32                                                                            & 194824                                                                              & $2.31\times10^{-2}$                                                                          \\ \hline
\multirow{4}{*}{12}         & 30               & 0.0700                              & 0.01757                              & \multicolumn{1}{l}{2.1653}                                       & 1370.67                                                                            & 37193                                                                               & $1.84\times10^{-3}$                                                                          \\
                            & 20               & 0.0570                              & 0.00775                              & 4.1027                                                           & \multicolumn{1}{r}{3331.26}                                                        & \multicolumn{1}{c}{115976}                                                          & $1.03\times10^{-3}$                                                      \\
                            & 15               & 0.2450                              & 0.04299                              & \multicolumn{1}{l}{1.2186}                                       & 4326.84                                                                            & 169276                                                                              & $4.69\times10^{-3}$                                                                          \\
                            & 10               & 0.2917                              & 0.09241                              & \multicolumn{1}{l}{1.2286}                                       & 2956.41                                                                            & 103524                                                                              & $2.40\times10^{-2}$                                                                          \\ \hline
\multirow{4}{*}{\textbf{9}} & 30               & 0.0566                              & 0.00994                              & 0.7713                                                           & \multicolumn{1}{c}{2600.01}                                                        & \multicolumn{1}{c}{102071}                                                          & $1.02\times10^{-3}$                                                                          \\
                            & \textbf{20}      & \multicolumn{1}{c}{\textbf{0.0562}} & \multicolumn{1}{c}{\textbf{0.01428}} & \textbf{1.2374}                                                  & \multicolumn{1}{c}{\textbf{2684.33}}                                               & \multicolumn{1}{c}{\textbf{101431}}                                                 & $\mathbf{9.56\times10^{-4}}$                                                      \\
                            & 15               & \multicolumn{1}{c}{0.0606}          & \multicolumn{1}{c}{0.10383}          & 1.9813                                                           & \multicolumn{1}{c}{3242.01}                                                        & \multicolumn{1}{c}{135796}                                                          & $1.06\times10^{-3}$                                                                          \\
                            & 10               & \multicolumn{1}{c}{0.3525}          & \multicolumn{1}{c}{0.10383}          & 0.8183                                                           & \multicolumn{1}{c}{4257.71}                                                        & \multicolumn{1}{c}{199267}                                                          & $1.83\times10^{-2}$                                                                          \\ \hline
\multirow{4}{*}{4}          & 30               & 0.0608                              & 0.01168                              & \multicolumn{1}{l}{0.7918}                                       & 2546.29                                                                            & 128028                                                                              & $1.42 \times10^{-3}$                                                                         \\
                            & 20               & \multicolumn{1}{c}{0.0690}          & \multicolumn{1}{c}{0.02296}          & 1.3998                                                           & \multicolumn{1}{c}{3719.84}                                                        & \multicolumn{1}{c}{199102}                                                          & $7.57\times10^{-3}$                                                       \\
                            & 15               & 0.0746                              & 0.02078                              & \multicolumn{1}{l}{0.5479}                                       & 3556.85                                                                            & 190957                                                                              & $6.73\times10^{-3}$                                                                          \\
                            & 10               & 0.0877                              & 0.03577                              & \multicolumn{1}{l}{0.5476}                                       & 3468.09                                                                            & 198833                                                                              & $1.62\times10^{-2}$                                                                          \\ \hline
\end{tabular}%
}
\end{table}

The experiments investigate the influence of the neural network architecture on state estimation performance, with particular emphasis on network depth, number of neurons per layer,  the choice of activation function, and the sensitivity of loss functions. Our analysis is guided by the need to balance model complexity, training efficiency, and generalization capability \cite{kuri2014best, oh2021optimal}.

Table~\ref{tab:performance_metrics} summarizes the performance of the observer for architectures. The results indicate that:
The 9-20 architecture delivered the highest accuracy among the tested models, as evidenced by the lowest error metrics in the provided table. In particular, it achieved the lowest RMSE and lowest MAE of all architectures compared.  The 9-20 model’s superior performance on both metrics indicates not only a low average error but also a low incidence of large outlier errors. This suggests that the model’s predictions are consistently close to the true values. For example, if we compare it to a smaller architecture (e.g., with fewer hidden neurons), the 9-20 model's RMSE is noticeably lower (indicating higher accuracy), and its MAE is similarly reduced, implying improvements in both overall predictive precision and reliability of individual predictions. Such an improvement in both metrics means the 9-20 architecture predicts more accurately and with fewer large deviations than other architectures.

 The 9-20 architecture also attained the lowest final loss during training. In the provided table, the "best loss value" for the 9-20 network is the smallest among all candidates, reinforcing that this model fit the training data most closely.  The fact that 9-20 achieved the best loss without overfitting (as supported by its low RMSE/MAE on test data) indicates an effective learning of the underlying data patterns. In contrast, other architectures showed higher loss values. For instance, a simpler network might be a plateau at a higher loss, unable to reduce error further due to limited capacity, whereas an overly complex network might achieve a low training loss but not translate that to low validation error (overfitting).

In terms of inference speed. According to the provided metrics, the 9-20 model’s inference time (the time to evaluate the model on new data) is competitive. It is only marginally higher than that of simpler architectures and significantly lower than that of more complex ones. For example, if a very small network (with fewer neurons) achieves a slightly faster inference per sample, it does so at the cost of accuracy; the 9-20 model, however, keeps inference time low enough to be practical (on the order of milliseconds per sample, as indicated in the table) while dramatically improving accuracy. Compared to a larger network architecture (with more neurons or additional layers), the 9-20 model is much faster during inference  a larger model might take notably longer to run, which can be prohibitive for real-time use or large scale data processing. Thus, the chosen model provides quick predictions, suitable for time sensitive applications, without the latency overhead that typically comes with very deep or wide networks.

The comprehensive evaluation justifies that the 9-20 architecture is the optimal choice for the task, as it balances accuracy, efficiency, and reliability more effectively than other designs.  The decision to select the 9-20 architecture is thus well-supported by the metrics and observations discussed above, aligning with best practices in model selection to meet both technical and practical requirements \cite{deng2023effect}.
\subsubsection{Activation function}
\begin{table}[ht]
\centering
\caption{Performance Metrics of the Adaptive PINN-based Observer for Different Activation Functions.}
\label{tab:activation_metrics}
\footnotesize
\begin{tabular}{lcccccl}
\hline
\textbf{Activation} & \textbf{RMSE} & \textbf{MAE} & \textbf{\begin{tabular}[c]{@{}c@{}}I- Time\\ (ms)\end{tabular}} & \textbf{\begin{tabular}[c]{@{}c@{}}T- Time\\ (s)\end{tabular}} & \textbf{\begin{tabular}[c]{@{}c@{}}Conv.\\ Iter.\end{tabular}} & \multicolumn{1}{c}{\textbf{\begin{tabular}[c]{@{}c@{}}Best\\ Loss\end{tabular}}} \\ \hline
Relu                & 0.9889       & 0.78610      & 1.2794                                                         & 1134.26                                                        & 31205         & $9.79\times10^{-1}$ \\
Sigmoid             & 1.4721       & 1.31830      & 1.2929                                                         & 702.45                                                         & 12923         & $2.81\times10^{0}$ \\
\textbf{Tanh}       & \textbf{0.0535}      & 0.00954      & \textbf{1.2374}                                                    & \textbf{3365.85}                                                & \textbf{135796}        & \textbf{$9.74\times10^{-4}$} \\
Sine                & 0.0561       & \textbf{0.00724}      & 1.3587                                                         & 4087.03                                                        & 166457        & $\mathbf{6.96\times10^{-4}}$ \\ \hline
\end{tabular}
\end{table}

Table~\ref{tab:activation_metrics} presents a comparative analysis of different activation functions. The observations are as follows: \begin{itemize} \item \textbf{ReLU and Sigmoid:} Both these functions resulted in considerably higher RMSE and loss values, reflecting poor performance in approximating the dynamics. \item \textbf{Tanh:} Achieved the lowest RMSE (0.0535) and an excellent loss metric, along with a moderate number of convergence iterations. Its smooth, bounded nonlinearity is especially effective in capturing the intricacies of the reverse Duffing oscillator. \item \textbf{Sine:} Offered performance comparable to Tanh; however, Tanh is favored due to its well-understood training properties and greater stability in deep networks. \end{itemize}
These results clearly justify the selection of the Tanh activation function for the observer, given its superior performance and robustness.
\subsubsection{Weight Sensitivity}
Table~\ref{tab:weight_sensitivity} summarizes the performance of the observer under various loss weight configurations. In these experiments, the loss function is composed of three terms associated with the initial condition (\(w_0\)), the ODE residual (\(w_{Ode}\)), and the output measurement (\(w_y\)) and the baseline configuration uses equal weights (i.e., \(w_0 = w_{Ode} = w_y = 1.0\)). The performance is evaluated in terms of the best achieved loss, the number of convergence iterations, and the RMSE.

\begin{table}[ht]
\centering
\caption{Sensitivity of the loss function to weight variations for the adaptive PINN-based observer.}
\label{tab:weight_sensitivity}
\begin{tabular}{ccccccc}
\hline
\textbf{Case} & \(\mathbf{w_{0}}\)& \textbf{\(\mathbf{w_{\text{ode}}}\)} & \textbf{\(\mathbf{w_{y}}\)} & \textbf{Best Loss}           & \textbf{\begin{tabular}[c]{@{}c@{}}Conv.\\ Iter.\end{tabular}} & \textbf{RMSE}              \\ \hline
1 (Baseline)  & 1.0                & 1.0                  & 1.0              & \(9.56\times10^{-4}\)         & 101431                                                         & 0.0562                     \\
2             & 0.5                & 1.5                  & 1.0              & \(1.491\times10^{-3}\)        & 115976                                                         & 0.0640                     \\
3             & 1.5                & 0.5                  & 1.0              & \(7.41\times10^{-4}\)         & 159851                                                         & 0.0530                     \\
4             & 1.0                & 2.0                  & 1.0              & \(1.260\times10^{-3}\)        & 180451                                                         & 0.0579                     \\
5             & 2.0                & 1.0                  & 1.0              & \(1.272\times10^{-3}\)        & 182441                                                         & 0.0592                     \\
6             & 2.0                & 1.0                  & 0.5              & \(6.28\times10^{-4}\)         & 166549                                                         & 0.0599                     \\
7             & 2.0                & 1.5                  & 1.5              & \(1.711\times10^{-3}\)        & 161942                                                         & 0.0598                     \\ \hline
\end{tabular}
\end{table}

\paragraph{Baseline (Case 1):}  
The baseline configuration uses uniform weights (\(w_0 = w_{ode} = w_{y} = 1.0\)) and serves as the reference, achieving a best loss of \(9.56\times10^{-4}\), convergence in 101431 iterations, and an RMSE of 0.0562. This balanced configuration ensures that each loss component contributes equally to the overall training objective.

\paragraph{Case 2:}  
By reducing the weight for the initial condition (\(w_0 = 0.5\)) and increasing the weight for the ODE residual (\(w_{ode} = 1.5\)), the best loss worsens to \(1.491\times10^{-3}\) and the RMSE increases to 0.0640, with a moderate increase in convergence iterations. This suggests that de-emphasizing the initial condition degrades the model's ability to capture the correct starting state, thereby impairing overall accuracy.

\paragraph{Case 3:}  
Increasing the weight for the initial condition (\(w_0 = 1.5\)) while reducing the ODE residual weight (\(w_{ode} = 0.5\)) yields an improved best loss of \(7.41\times10^{-4}\) and a lower RMSE of 0.0530, albeit at the cost of a higher number of convergence iterations (159851). Emphasizing the initial condition appears to enhance prediction accuracy, though it may require additional training iterations to fully converge.

\paragraph{Case 4:}  
With an even higher emphasis on the ODE residual (\(w_{ode} = 2.0\)) and baseline weights for the other components, the best loss increases to \(1.260\times10^{-3}\) and convergence iterations rise to 180451, while the RMSE remains similar to the baseline (0.0579). This indicates that overemphasizing the ODE residual can slow convergence without yielding substantial improvements in prediction accuracy.

\paragraph{Case 5:}  
Conversely, when the initial condition weight is increased to \(w_0 = 2.0\) (with baseline values for the other terms), the best loss becomes \(1.272\times10^{-3}\), convergence iterations are highest at 182441, and the RMSE increases to 0.0592. This over-weighting of the initial condition may lead to an imbalance that negatively affects overall performance.

\paragraph{Case 6:}  
Reducing the output measurement weight to \(w_{y} = 0.5\) while setting \(w_0 = 2.0\) and \(w_{ode} = 1.0\) results in the lowest best loss (\(6.28\times10^{-4}\)); however, the RMSE (0.0599) does not show a corresponding improvement relative to the baseline. This indicates that while the overall loss is minimized, underweighting the measurement component might compromise the model’s ability to accurately track the observed output.

Finally, increasing the weights for both the ODE residual and the output measurement (\(w_{ode} = 1.5\) and \(w_{y} = 1.5\)) with \(w_0 = 2.0\) leads to the highest best loss (\(1.711\times10^{-3}\)) and an RMSE of 0.0598. This combination of high weights does not yield performance benefits; rather, it appears to degrade the overall effectiveness of the training process.

The results demonstrate that a balanced weighting strategy is crucial for achieving optimal performance. The baseline configuration provides a robust reference, while moderate adjustments such as in Case 3 can slightly improve RMSE, although often at the expense of longer convergence times. Overemphasizing any single component (as in Cases 4, 5, and 7) tends to disrupt the equilibrium, resulting in either higher loss values or degraded accuracy. Notably, while Case 6 achieves the lowest best loss, the corresponding RMSE does not improve, suggesting that the minimum loss is not always directly correlated with better predictive performance if one of the key loss components is underweighted.

In summary, these experiments underscore the importance of carefully balancing the loss weights to simultaneously minimize training loss, reduce convergence iterations, and achieve low RMSE. The findings provide valuable insights into the sensitivity of each loss component, guiding the design of more robust and accurate observer models.

\subsection{Comparison with Existing Methods}
In this section, we evaluate the Adaptive PINN  based  observer's  performance against the following baselines:
\begin{itemize}
    \item \textbf{Supervised  NN} \cite{ramos2020numerical}:  This method employs a supervised regression approach for continuous-time systems. Simulated trajectories from both the nonlinear system and its linear observer are used to train a neural network that approximates the transformation (T) and its inverse mapping (T*).
    \item  \textbf{Unsupervised  AE} \cite{peralez2021deep}: Designed for discrete-time systems, this method utilizes a deep autoencoder architecture in an unsupervised manner. The network learns both the transformation (T) and its inverse by incorporating the observer dynamics into the loss function.
    \item  \textbf{Supervised PINNs} \cite{niazi2023learning}: This approach enhances observer design by incorporating physics-informed constraints (i.e., the ordinary differential equation governing T) into the supervised learning framework. This improves generalization and robustness against model uncertainties and sensor noise.
\end{itemize}
The key differences between these methods lie in the system 
setting (continuous versus discrete time), the learning strategy (pure supervised 
regression, unsupervised, and supervised PINNs), and the integration of physical knowledge, which in third method  helps to 
reduce overfitting and improve performance. Together, these methods represent an evolution from conventional regression-based approaches toward more robust, constraint-driven designs that better capture the underlying system dynamics.

Although these methods demonstrate promising performance in estimating nonlinear system states, estimating the linear transformation to recast a nonlinear system into a linear form presents several challenges. First, the transformation $T$ is defined implicitly by an ordinary differential equation, which rarely has an explicit solution. This necessitates reliance on numerical or data-driven approximations that may only capture local behavior. Second, even minor estimation errors in $T$ can be significantly amplified when computing its left inverse $T^*$, particularly in regions where $T$ is nearly singular or exhibits low sensitivity. This can lead to substantial observer inaccuracies. Third, the quality and density of training data are critical. If the sampled data does not adequately cover the state space, the learned transformation may overfit to specific regions and fail to generalize. Lastly, the approach relies on strong properties such as backward distinguishability and sufficient observability of the system, which may not hold for many practical nonlinear systems, thereby limiting the method’s applicability.

Our method not only estimates the state but also adaptively learns a state dependent gain so that the estimated state converges to the true state using only the system output. The key idea is to embed the observer structure into the PINN framework and enforce convergence via physics–based loss functions. 

\begin{figure}[!t]
    \centering
    \subfloat[Comparison of predicted and actual state trajectories of  Reverse Duffing Oscillator. ]{
        \includegraphics[width=0.6\linewidth]{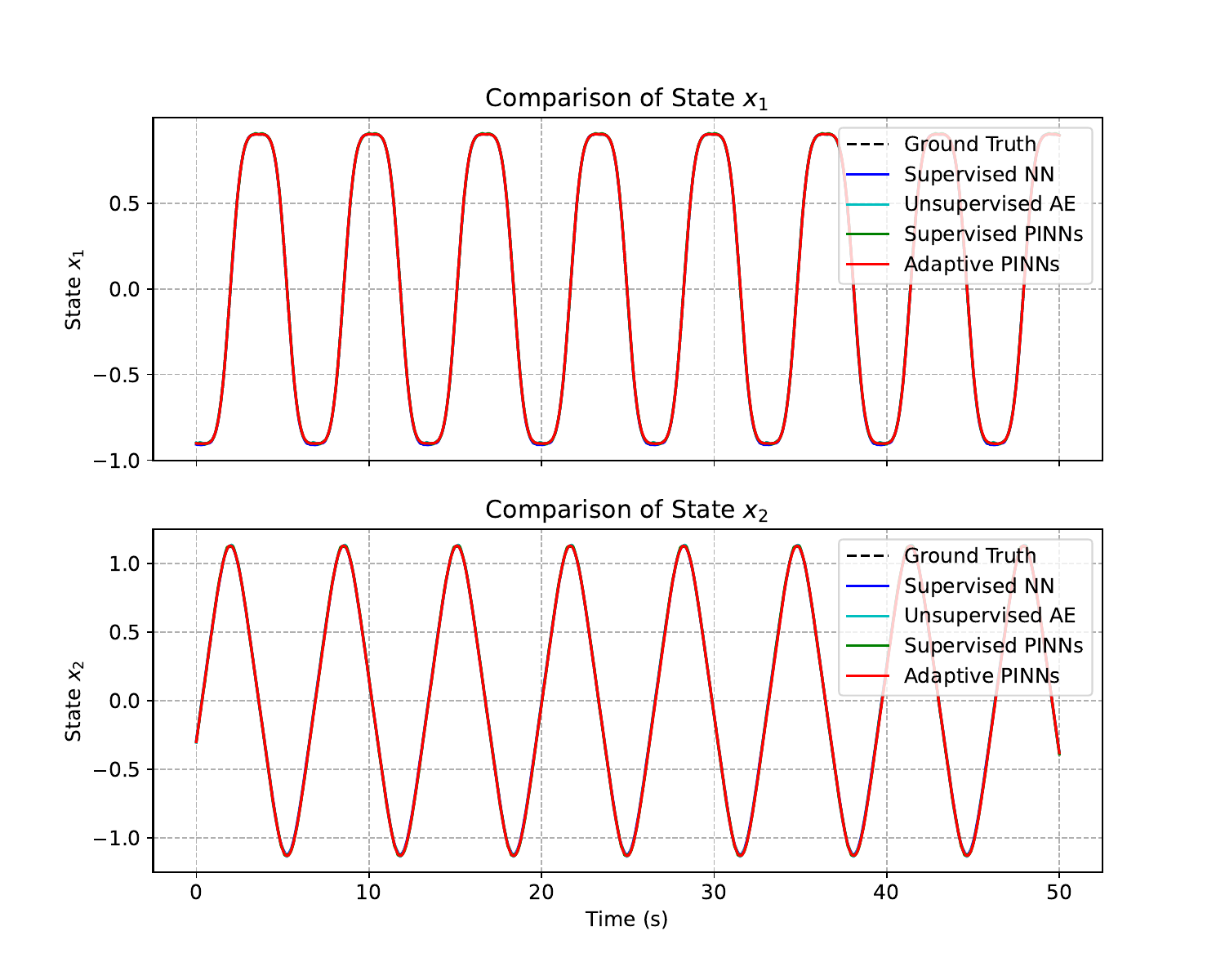}
        \label{f2a1}
    }\\
    \subfloat[Comparison of error metrics (MSE, RMSE, MAE, and SMAPE) across four methods.]{
        \includegraphics[width=0.6\linewidth]{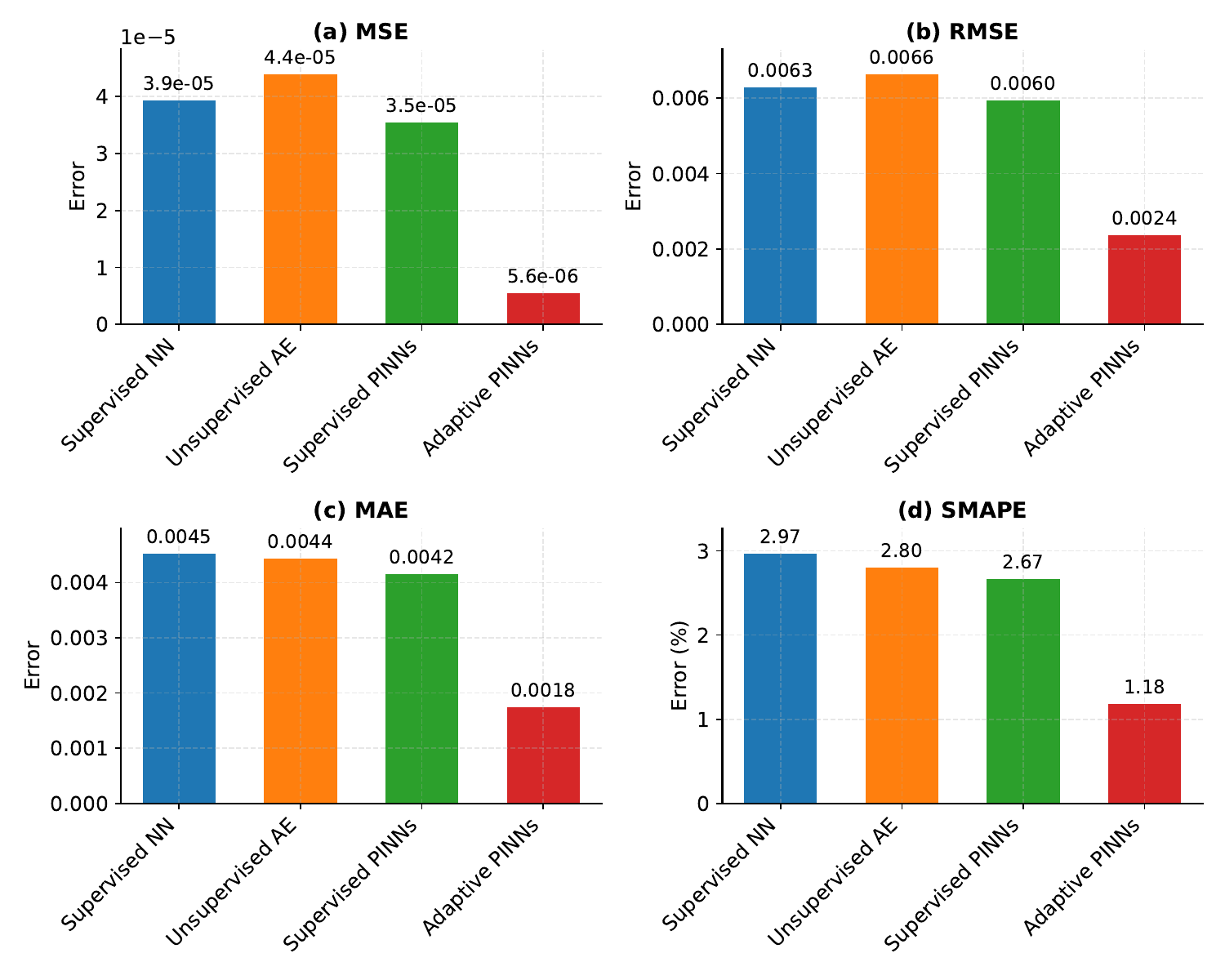}
        \label{f2b1}
    }
    \caption{(a) Comparison of predicted and actual state trajectories, and (b) Comparison of our method with other approaches using various error metrics.}
    \label{f2-1}
\end{figure}
The comparison is conducted on the  Reverse Duffing Oscillator~(\ref{reverse}). The Results are illustared in the Figure~\ref{f2-1}. 
The results clearly indicate that the Adaptive PINN observer offers a marked improvement over the other methods.  the Adaptive PINN achieved an MSE of \(5.56 \times 10^{-6}\), which is roughly an order of magnitude lower than that of the Supervised PINNs (\(3.54 \times 10^{-5}\)), Supervised NN (\(3.94 \times 10^{-5}\)), and Unsupervised AE (\(4.39 \times 10^{-5}\)). Similar improvements are observed in the RMSE values, with the Adaptive PINN recording 0.00236 compared to 0.00595, 0.00628, and 0.00663, respectively. The MAE and SMAPE metrics follow the same trend Adaptive PINN achieves a MAE of 0.00175 and a SMAPE of $1.18\%$, which are substantially lower than the corresponding error values from the other approaches. These performance gains suggest that the Adaptive PINN is not only more accurate but also more robust in state estimation, significantly reducing both absolute and relative errors. 

Next experiments will extend this framework to
 more complex dynamical scenarios, further validating its scalability and robustness.
\subsection{Experimental Setup and Neural Network Architecture}
For all applications, we designed a comprehensive neural network architecture tailored to the observer design task. The architecture comprises $9$ hidden layers, each containing $20$ neurons, which provides sufficient capacity to extract intricate patterns from the available data. We employ the hyperbolic tangent (\( \tanh \)) activation function to ensure smooth nonlinearity modeling and to preserve stability through its boundedness and 1-Lipschitz property. The network is trained using the Adam optimizer \cite{kingma2014adam} with an initial learning rate of \(0.001\) to balance rapid convergence with the avoidance of local minima. The dataset is partitioned with 60\% of the samples allocated to training and the remaining 40\% reserved for testing. Training was accelerated using a Tesla T4 GPU, which significantly reduced the overall training time and enabled extensive hyperparameter exploration.
\begin{ex}
Consider a detailed fifth-order, two-phase nonlinear model of an induction motor. This model, widely used in control design \cite{boutayeb1999strong, marino1993adaptive}, relies solely on stator current measurements to estimate the rotor's magnetic flux and rotational speed. The system is reformulated as:
\begin{equation}
\left\{
\begin{aligned}
\dot{x}_{1} & = -\gamma x_{1}+\frac{K}{T_r} x_{3}+K p x_{5} x_{4}+\frac{1}{\sigma L_s} u_{1}, \\
\dot{x}_{2} & = -\gamma x_{2}-K p x_{5} x_{3}+\frac{K}{T_r} x_{4}+\frac{1}{\sigma L_s} u_{2}, \\
\dot{x}_{3} & = \frac{M}{T_r} x_{1}-\frac{1}{T_r} x_{3}-p x_{5} x_{4}, \\
\dot{x}_{4} & = \frac{M}{T_r} x_{2}+p x_{5} x_{3}-\frac{1}{T_r} x_{4}, \\
\dot{x}_{5} & = \frac{p M}{J L_r}\left(x_{3} x_{2}-x_{4} x_{1}\right)-\frac{T_L}{J}, \\
y_{1} & = x_{1}, \quad y_{2} = x_{2}.
\end{aligned}
\right.
\end{equation}
Here, \(x=[x_1~~ x_2~~ x_3~~ x_4~~ x_5]^T\) includes the stator currents, rotor fluxes, and angular speed, while \(u=[u_1~~ u_2]^T\) is the stator voltage control vector. The rotor time constant, leakage factor, and other parameters are defined as:
\[
T_r=\frac{L_r}{R_r}, \quad \sigma=1-\frac{M^2}{L_sL_r}, \quad K=\frac{M}{\sigma L_sL_r}, \quad \gamma=\frac{R_s}{\sigma L_s}+\frac{R_r M^2}{\sigma L_sL_r^2}.
\]
\begin{table}[htbp]
\centering
\begin{tabular}{lc}
\hline
Parameters & Values                  \\ \hline
$R_s$      & $0.18 \Omega$           \\
$R_r$      & $0.15 \Omega$           \\
$M$        & $0.068 \mathrm{H}$      \\
$L_s$      & $0.0699 \mathrm{H}$      \\
$L_r$      & $0.0699 \mathrm{H}$     \\
$J$        & $0.0586 \mathrm{kg}\cdot m^2$ \\
$T_L$      & $10 \mathrm{Nm}$        \\
$p$        & 1        \\ \hline              
\end{tabular}
\caption{The simulation model parameters of induction motor system. }
\label{t1}
\end{table}
Table~\ref{t1} lists the simulation parameters. The dataset is obtained from sensors measuring \(x_1\) and \(x_2\) over the time interval \(\Omega=[0,20]\). Initial conditions are set as:
\[
x_0 =[1~~ 0~~ 2~~3~~0]^T,\quad \hat{x}_0=[2~~ 1~~ 0~~2~~0]^T.
\]
As shown in Figures~\ref{f2a} and~\ref{f2b}, the network successfully estimates the unmeasured state components \(x_3\), \(x_4\), and \(x_5\) based solely on \(x_1\) and \(x_2\), and the estimation error converges as \(t\) increases.
\begin{figure}[!t]
    \centering
    \subfloat[Predicted and actual state trajectories for the induction motor system.]{
        \includegraphics[width=0.5\linewidth]{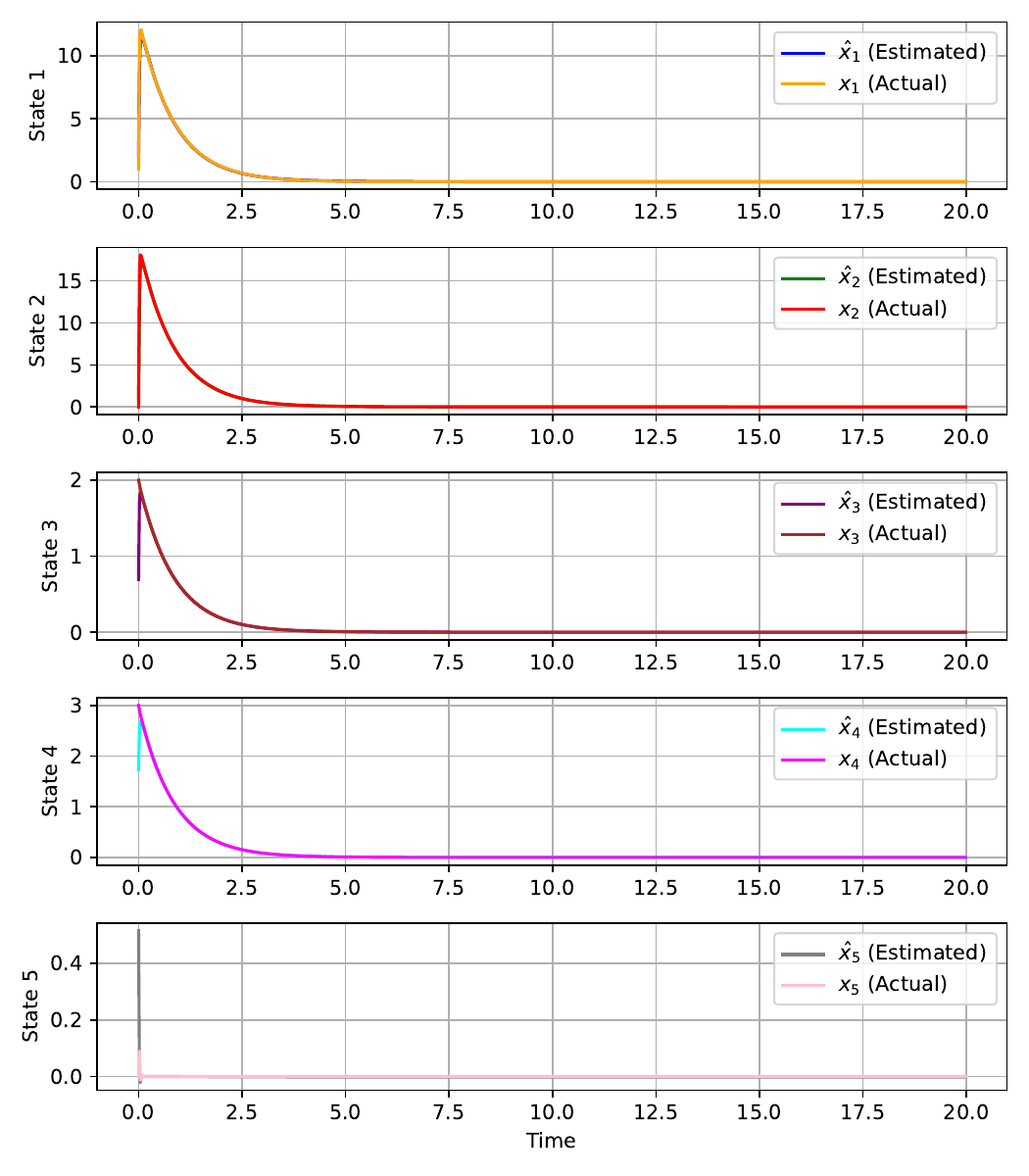}
        \label{f2a}
    }
    \subfloat[Time evolution of the estimation error for the induction motor system.]{
        \includegraphics[width=0.5\linewidth]{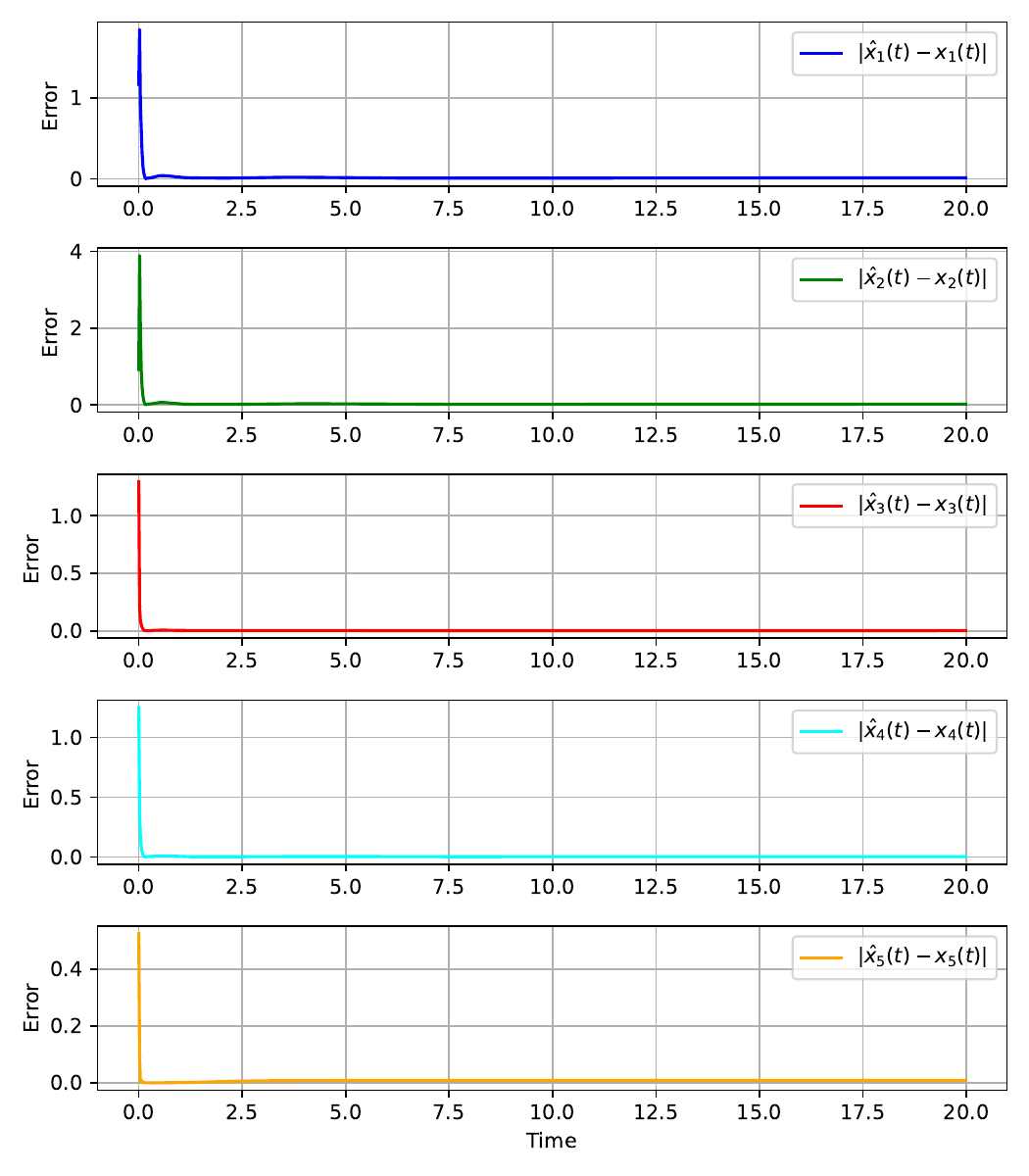}
        \label{f2b}
    }
    \caption{(a) Comparison of predicted and actual state trajectories, and (b) corresponding estimation errors for the induction motor system.}
\end{figure}
\end{ex}
\begin{ex}
A harmonic oscillator with an unknown frequency is fundamental in many mechanical applications \cite{sanfelice2015convergence, praly2006new}. The system is described by:
\begin{equation}
\left\{
\begin{aligned}
\dot{x}_1 &= x_2, \\
\dot{x}_2 &= -x_3\,x_1, \\
\dot{x}_3 &= 0, \\
y &= x_1.
\end{aligned}
\right.
\end{equation}
\begin{figure}[!t]
    \centering
    \subfloat[Estimated and actual states of the harmonic oscillator system.]{
        \includegraphics[width=0.5\linewidth]{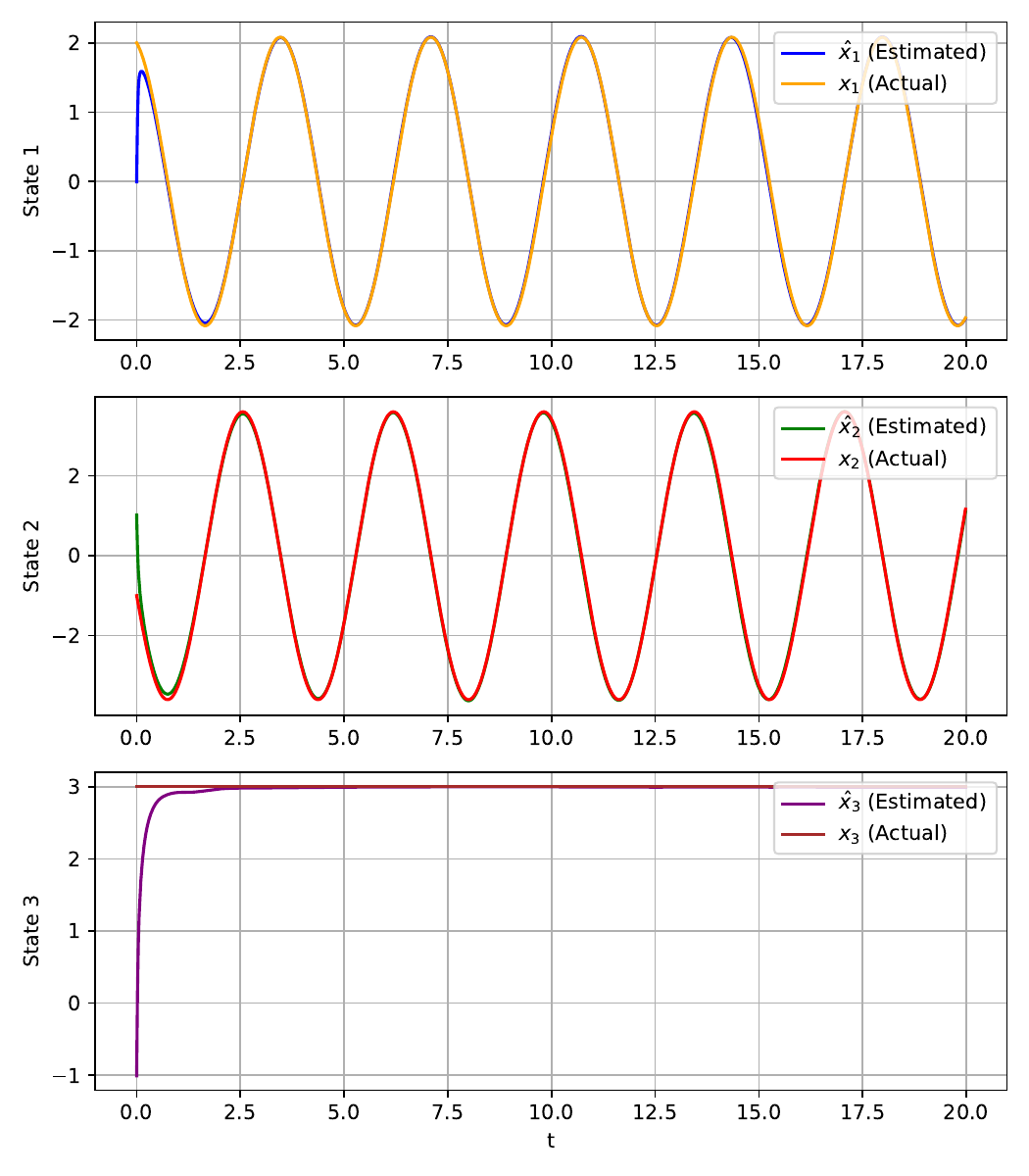}
        \label{f4a}
    }
    \subfloat[Time evolution of the estimation error for the harmonic oscillator system.]{
        \includegraphics[width=0.5\linewidth]{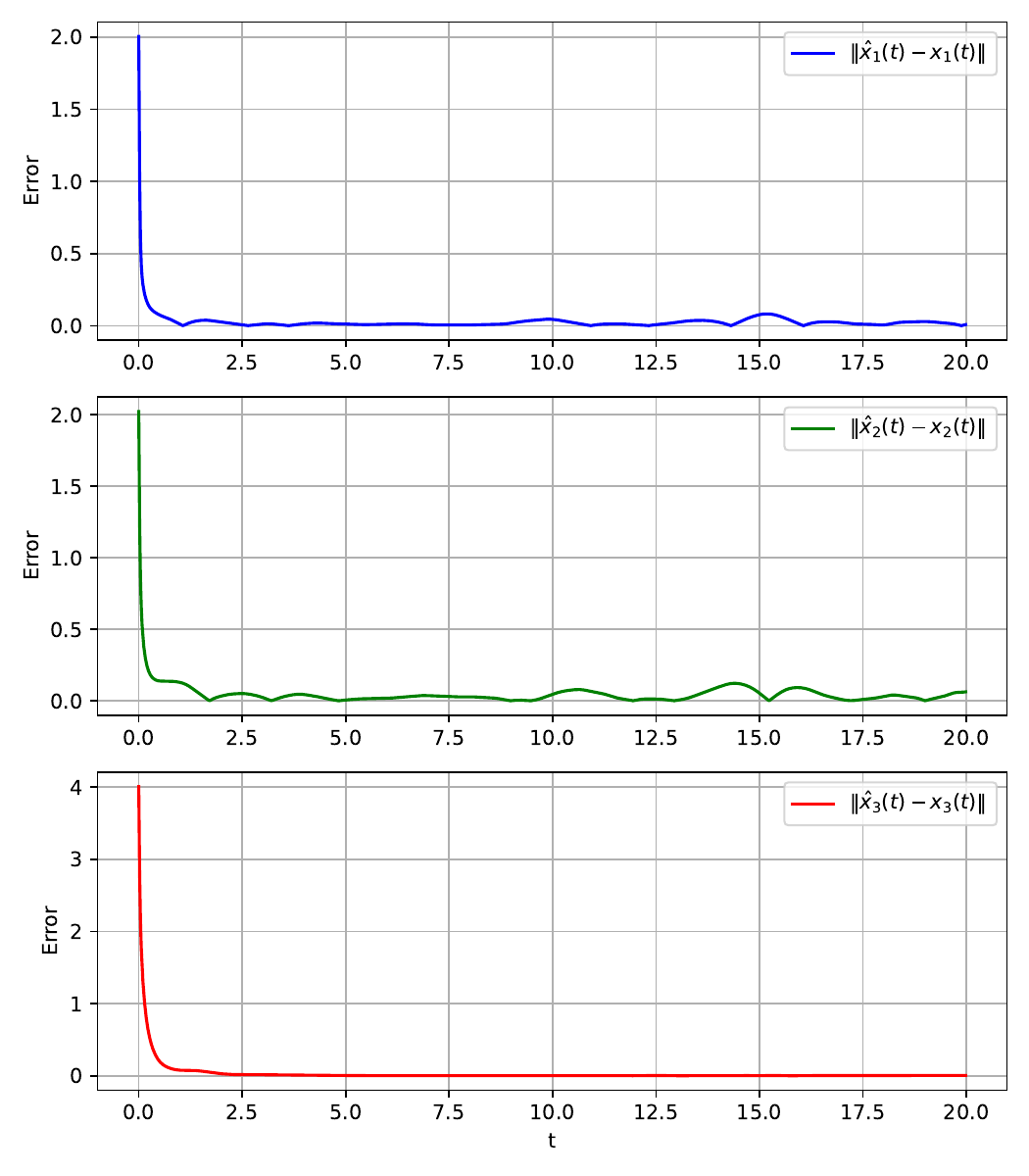}
        \label{f4b}
    }
    \caption{(a) Comparison of estimated and actual states, and (b) corresponding estimation errors for the harmonic oscillator system.}
    \label{f4}
\end{figure}
Here, the measured variable is \(x_1\). The initial conditions are:
\[
x_0=[0 \quad 1 \quad 3]^T,\quad \hat{x}_0=[0 \quad 1 \quad -1]^T.
\]
Figures~\ref{f4a} and~\ref{f4b} illustrate that the predicted state trajectories nearly overlap with the actual ones. In particular, the constant behavior of \(x_3\) is accurately captured and the convergence of the state estimate is clearly demonstrated by the low estimation errors.
\end{ex}
\begin{ex}\label{ex3}
Consider the academic system:
\begin{equation}
\left\{
\begin{aligned}
\dot{x}_1 &= x_2 \sqrt{1+x_1^2}, \\
\dot{x}_2 &= -\frac{x_1}{\sqrt{1+x_1^2}} x_2^2, \\
y &= x_1.
\end{aligned}
\right.
\end{equation}
Detectability for this system is addressed in \cite{sanfelice2011convergence} using the Riemann metric. Figures~\ref{f6a} and~\ref{f6b} show that the predicted state trajectories \( \hat{x}_1 \) and \( \hat{x}_2 \) converge to the actual states \( x_1 \) and \( x_2 \), respectively.
\begin{figure}[!t]
    \centering
    \subfloat[Predicted and actual state vectors for academic example \ref{ex3}.]{
        \includegraphics[width=0.5\linewidth]{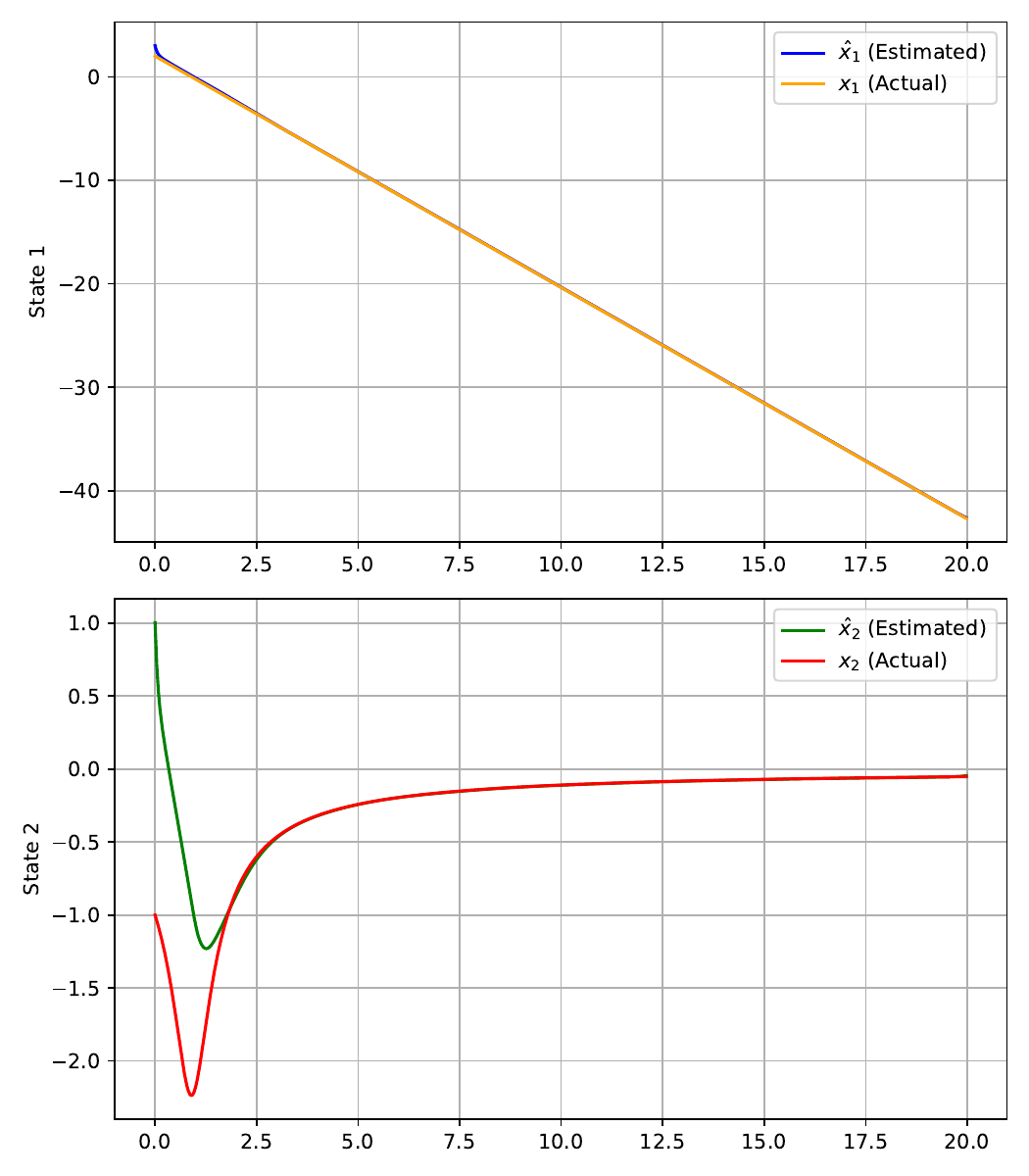}
        \label{f6a}
    }
    \subfloat[Estimated errors for academic example \ref{ex3}.]{
        \includegraphics[width=0.556\linewidth]{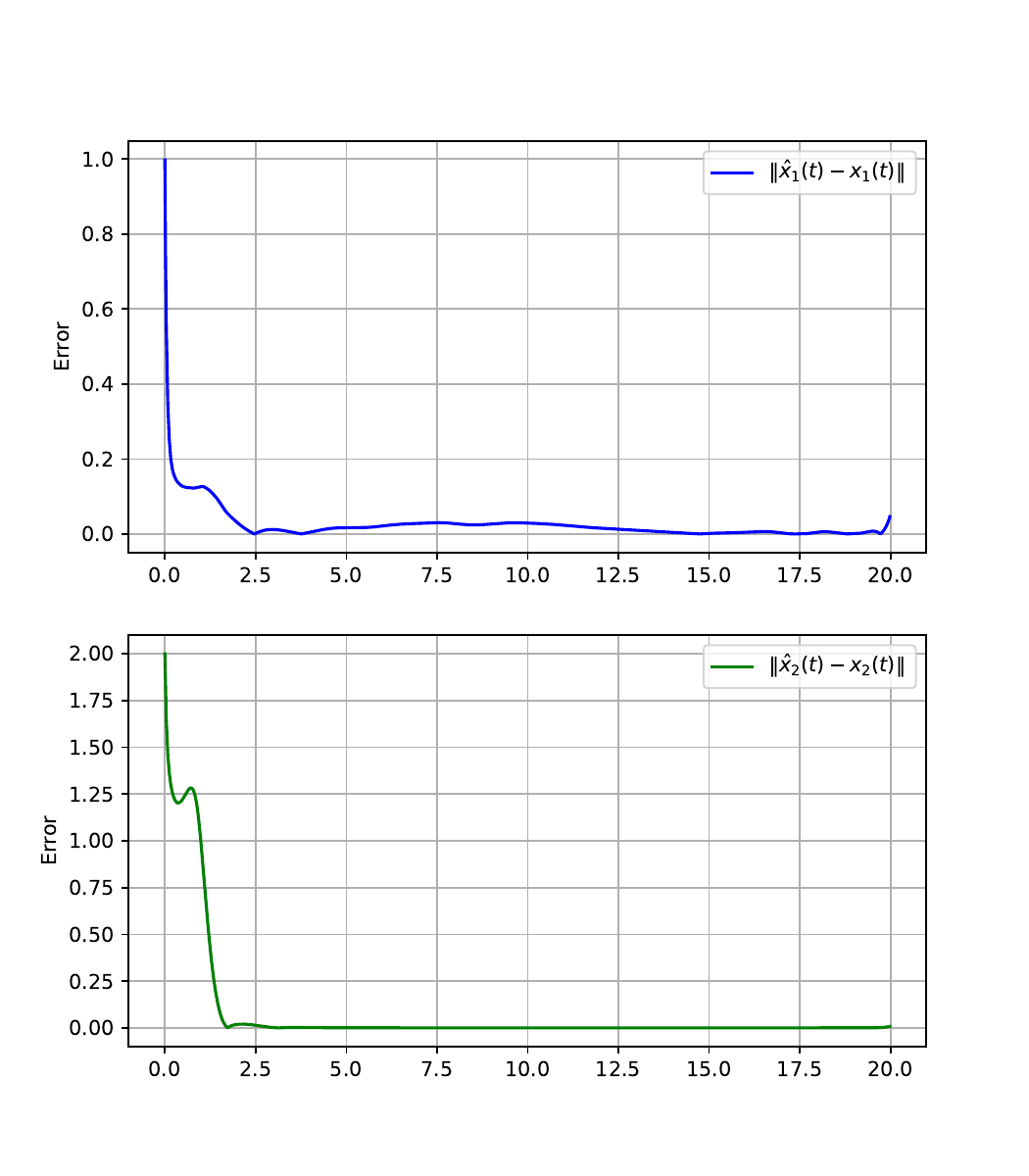}
        \label{f6b}
    }
    \caption{(a) Predicted and actual state vectors and (b) corresponding estimation errors for academic example \ref{ex3}.}
    \label{f6_combined}
\end{figure}

\end{ex}
\begin{ex}\label{ex4}
To further challenge the observer design, we consider a modified version of the academic system:
\begin{equation}
\left\{
\begin{aligned}
\dot{x}_1 &= x_2 \sqrt{1+x_2^2}, \\
\dot{x}_2 &= -\frac{x_1}{\sqrt{1+x_2^2}} x_2^2, \\
y &= x_1.
\end{aligned}
\right.
\end{equation}
Although the observability in this case is non-trivial due to the altered nonlinearity, Figures~\ref{f8a} and~\ref{f8b} confirm that the neural network still estimates the state vectors accurately. 
\begin{figure}[!t]
    \centering
    \subfloat[Predicted and actual state vectors for academic example \ref{ex4}.]{
        \includegraphics[width=0.5\linewidth]{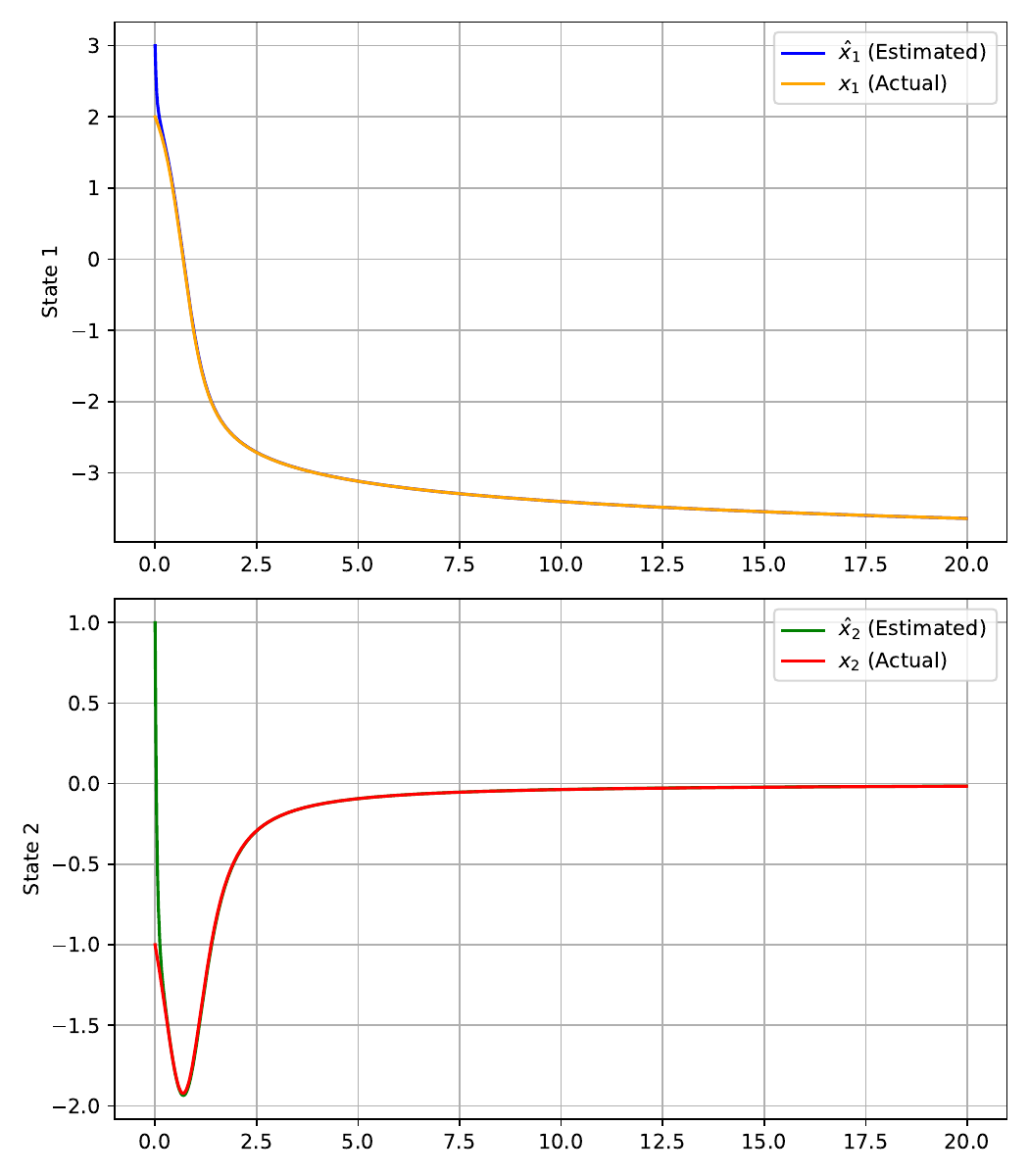}
        \label{f8a}
    }
    \subfloat[Estimated errors for academic example \ref{ex4}.]{
        \includegraphics[width=0.556\linewidth]{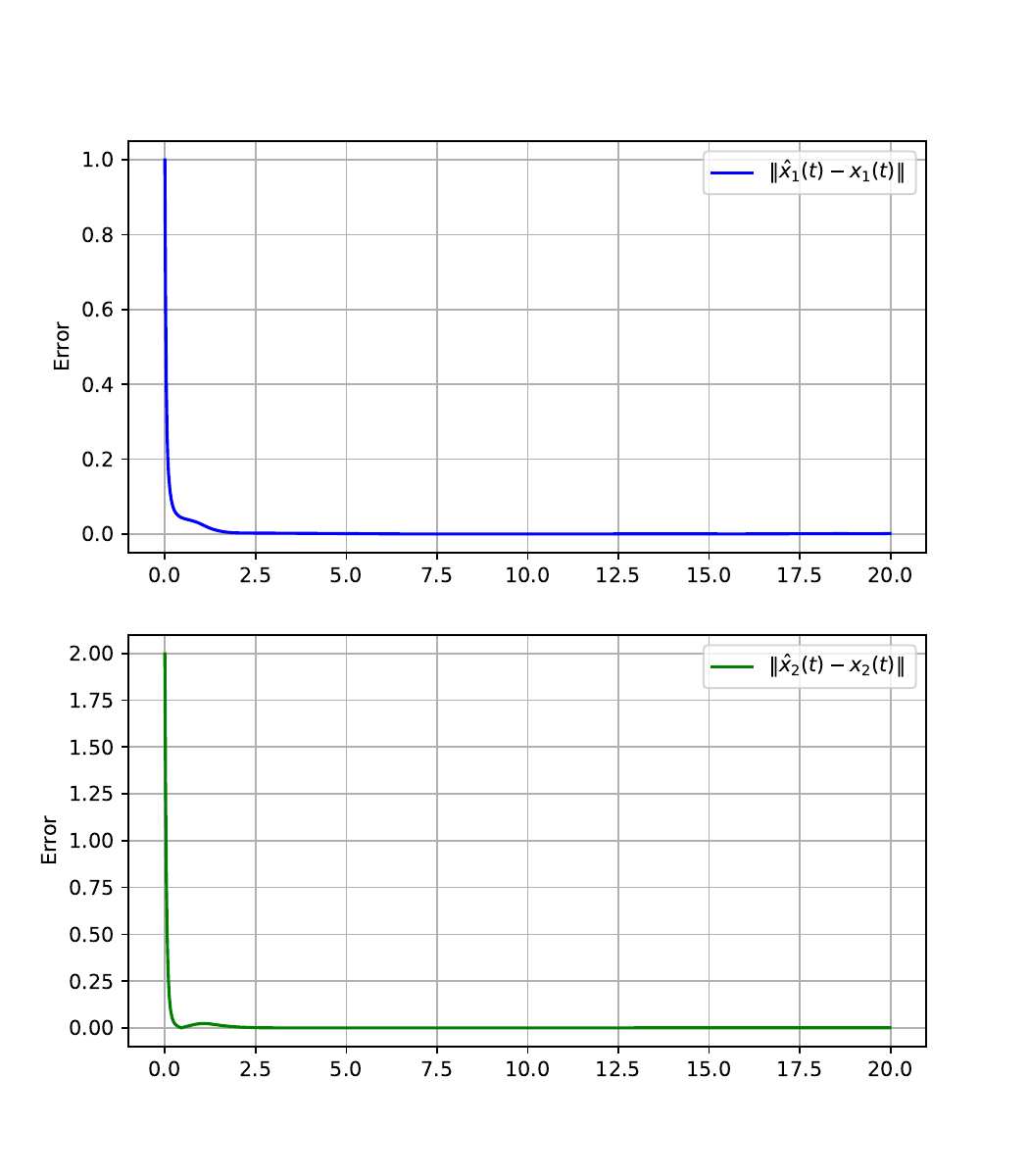}
        \label{f8b}
    }
    \caption{(a) Predicted and actual state vectors and (b) corresponding estimation errors for academic example \ref{ex4}.}
    \label{f8_combined}
\end{figure}

\end{ex}
\begin{ex}
Consider the dynamics of a rigid body in motion, modeled by:
\begin{equation}
\left\{
\begin{aligned}
\dot{x}_1 &= a_1\,x_2\,x_3, \\
\dot{x}_2 &= a_2\,x_1\,x_3, \\
\dot{x}_3 &= a_3\,x_1\,x_2, \\
y &= x_1,
\end{aligned}
\right.
\end{equation}
where \(x=[x_1 \; x_2 \; x_3]^T\) represents the angular velocity components along the body-fixed axes. The parameters \(a_1\), \(a_2\), and \(a_3\) are given by:
\[
a_1=\frac{I_2-I_3}{I_1}, \quad a_2=\frac{I_3-I_1}{I_2}, \quad a_3=\frac{I_1-I_2}{I_3},
\]
with \(I_1\), \(I_2\), and \(I_3\) being the principal moments of inertia. Previous studies have addressed control and reduced-order observer design for such systems \cite{outbib1992stabilizability, boutayeb2000reduced, jammazi2021global}. However, estimating two state variables from a single measured state in continuous time remains challenging. Despite the nonlinear function \(f\) not being globally Lipschitz, our approach yields high estimation accuracy, as evidenced by Figures~\ref{f10a} and~\ref{f10b}.
\begin{figure}[!t]
    \centering
    \subfloat[Estimated and actual state vectors of the satellite motion system.]{
        \includegraphics[width=0.55\linewidth]{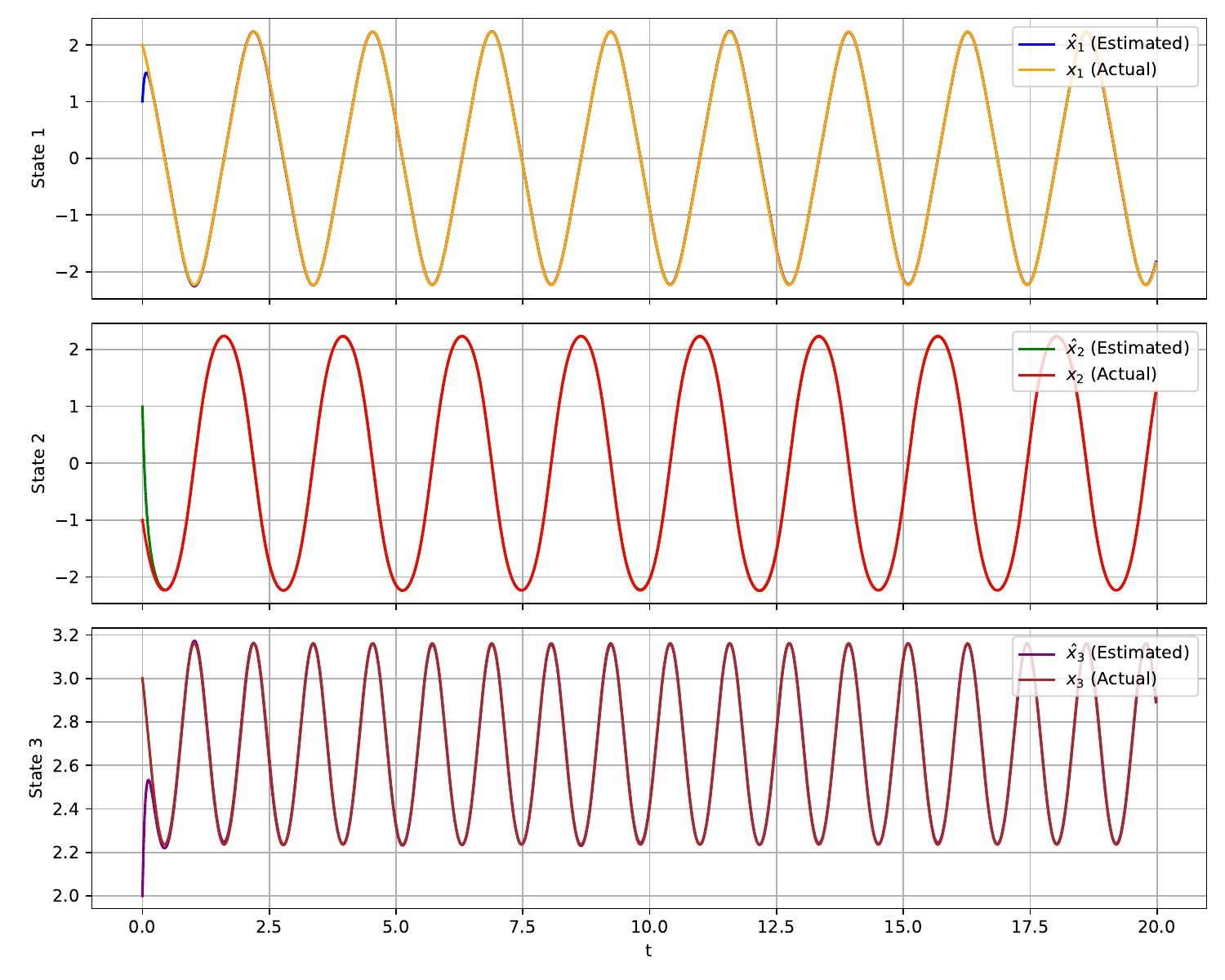}
        \label{f10a}
    }
    \subfloat[Estimated errors of the satellite motion system.]{
        \includegraphics[width=0.55\linewidth]{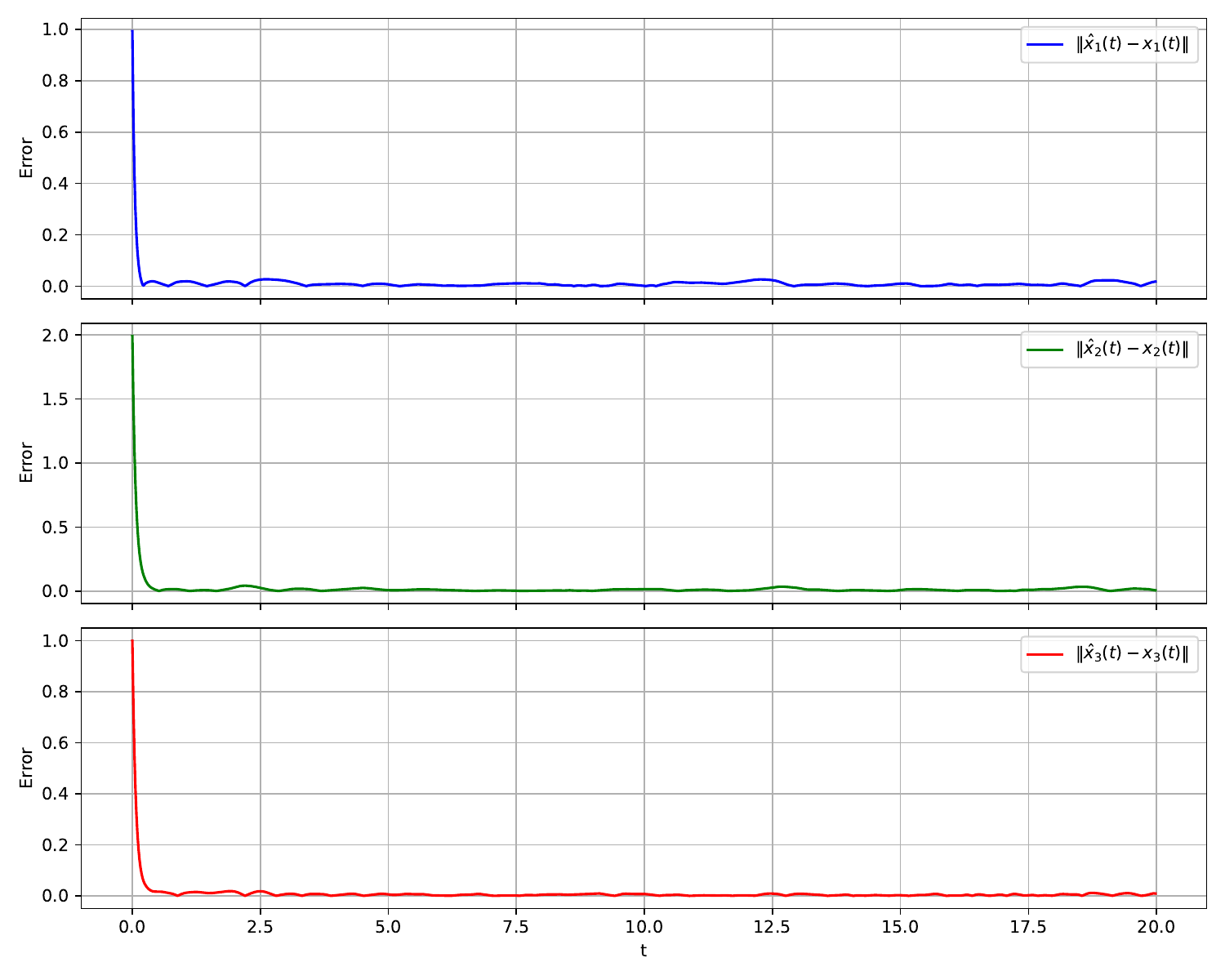}
        \label{f10b}
    }
    \caption{(a) Estimated and actual state vectors, and (b) corresponding estimation errors for the satellite motion system.}
    \label{f10_combined}
\end{figure}

\end{ex}
\subsection{Summary}
Across these examples, our adaptive PINN-based observer converges to the true system states even when only partial measurements are available. In the induction motor application, only two sensor measurements are used to estimate three unmeasured state variables, while in the harmonic oscillator and academic examples, the observer successfully captures the nonlinear dynamics and ensures convergence to the correct state. The rigid body example further demonstrates that our approach remains effective even when the system’s nonlinearity and observability pose significant challenges.

Overall, these applications validate the robustness, accuracy, and generalization capability of the proposed method in handling complex nonlinear system dynamics. The results underscore the method's potential for real-world applications in control and estimation, as well as its adaptability to different system complexities and measurement conditions.
\section{Conclusion}
This work introduced a novel Physics-Informed Neural Network-based observer (PINN-Obs) for state estimation in nonlinear dynamical systems. By integrating sensor data and leveraging the underlying physics, our approach achieves accurate state vector approximation without restrictive assumptions or coordinate transformations. Extensive validation in diverse applications including induction motors, satellite motion, and academic benchmark systems demonstrated the robustness, adaptability, and superior estimation accuracy of the method compared to existing approaches.

Future research may focus on enhancing the robustness of PINN-Obs against model uncertainties and external disturbances, particularly in the presence of noisy measurements. Investigating noise-aware training strategies, such as adversarial learning and probabilistic PINN formulations, could further improve resilience in real-world applications. Additionally, extending the framework to hybrid and stochastic systems, where process noise plays a significant role, presents a potential direction for exploration. Another promising direction involves real-time implementation and optimization, particularly for high-dimensional systems where computational efficiency is crucial. By addressing these challenges, PINN-Obs has the potential to become a standard tool for state estimation in complex and uncertain environments.
 \bibliographystyle{elsarticle-num} 

\begin{thebibliography}{10}
\expandafter\ifx\csname url\endcsname\relax
  \def\url#1{\texttt{#1}}\fi
\expandafter\ifx\csname urlprefix\endcsname\relax\def\urlprefix{URL }\fi
\expandafter\ifx\csname href\endcsname\relax
  \def\href#1#2{#2} \def\path#1{#1}\fi

\bibitem{niazi2023learning}
M.~U.~B. Niazi, J.~Cao, X.~Sun, A.~Das, K.~H. Johansson, Learning-based design of luenberger observers for autonomous nonlinear systems, in: 2023 American Control Conference (ACC), IEEE, 2023, pp. 3048--3055.

\bibitem{alvarez2024nonlinear}
H.~V. Alvarez, G.~Fabiani, N.~Kazantzis, I.~G. Kevrekidis, C.~Siettos, Nonlinear discrete-time observers with physics-informed neural networks, Chaos, Solitons \& Fractals 186 (2024) 115215.

\bibitem{marani2024unsupervised}
Y.~Marani, T.~Al-Naffouri, T.-M. Laleg-Kirati, et~al., Unsupervised physics-informed neural network-based nonlinear observer design for autonomous systems using contraction analysis, arXiv preprint arXiv:2411.09237 (2024).

\bibitem{de2024hybrid}
J.~de~Curt{\`o}, I.~de~Zarz{\`a}, Hybrid state estimation: Integrating physics-informed neural networks with adaptive ukf for dynamic systems, Electronics 13~(11) (2024) 2208.

\bibitem{ramos2020numerical}
L.~d.~C. Ramos, F.~Di~Meglio, V.~Morgenthaler, L.~F.~F. da~Silva, P.~Bernard, Numerical design of luenberger observers for nonlinear systems, in: 2020 59th IEEE Conference on Decision and Control (CDC), IEEE, 2020, pp. 5435--5442.

\bibitem{boutayeb1997convergence}
M.~Boutayeb, H.~Rafaralahy, M.~Darouach, Convergence analysis of the extended kalman filter used as an observer for nonlinear deterministic discrete-time systems, IEEE transactions on automatic control 42~(4) (1997) 581--586.

\bibitem{boutayeb1999strong}
M.~Boutayeb, D.~Aubry, A strong tracking extended kalman observer for nonlinear discrete-time systems, IEEE Transactions on Automatic Control 44~(8) (1999) 1550--1556.

\bibitem{song1992extended}
Y.~Song, J.~W. Grizzle, The extended kalman filter as a local asymptotic observer for nonlinear discrete-time systems, in: 1992 American control conference, IEEE, 1992, pp. 3365--3369.

\bibitem{raissi2019physics}
M.~Raissi, P.~Perdikaris, G.~E. Karniadakis, Physics-informed neural networks: A deep learning framework for solving forward and inverse problems involving nonlinear partial differential equations, Journal of Computational physics 378 (2019) 686--707.

\bibitem{arnold2021state}
F.~Arnold, R.~King, State--space modeling for control based on physics-informed neural networks, Engineering Applications of Artificial Intelligence 101 (2021) 104195.

\bibitem{delavari2023adaptive}
H.~Delavari, A.~Sharifi, Adaptive reinforcement learning interval type ii fuzzy fractional nonlinear observer and controller for a fuzzy model of a wind turbine, Engineering Applications of Artificial Intelligence 123 (2023) 106356.

\bibitem{varey2024physics}
J.~Varey, J.~D. Ruprecht, M.~Tierney, R.~Sullenberger, Physics-informed neural networks for satellite state estimation, in: 2024 IEEE Aerospace Conference, IEEE, 2024, pp. 1--8.

\bibitem{falas2023physics}
S.~Falas, M.~Asprou, C.~Konstantinou, M.~K. Michael, Physics-informed neural networks for accelerating power system state estimation, in: 2023 IEEE PES Innovative Smart Grid Technologies Europe (ISGT EUROPE), IEEE, 2023, pp. 1--5.

\bibitem{liu2024enhanced}
Y.~Liu, Y.~Bao, P.~Cheng, D.~Shen, G.~Chen, H.~Xu, Enhanced robot state estimation using physics-informed neural networks and multimodal proprioceptive data, in: Sensors and Systems for Space Applications XVII, Vol. 13062, SPIE, 2024, pp. 144--160.

\bibitem{stiasny2023physics}
J.~Stiasny, S.~Chatzivasileiadis, Physics-informed neural networks for time-domain simulations: Accuracy, computational cost, and flexibility, Electric Power Systems Research 224 (2023) 109748.

\bibitem{chen2018neural}
R.~T. Chen, Y.~Rubanova, J.~Bettencourt, D.~K. Duvenaud, Neural ordinary differential equations, Advances in neural information processing systems 31 (2018).

\bibitem{uccak2024adaptive}
K.~U{\c{c}}ak, G.~{\"O}. G{\"u}nel, Adaptive stable backstepping controller based on support vector regression for nonlinear systems, Engineering Applications of Artificial Intelligence 129 (2024) 107533.

\bibitem{shanbhag2025machine}
S.~Shanbhag, D.~E. Chang, Machine learning based state observer for discrete time systems evolving on lie groups, Engineering Applications of Artificial Intelligence 139 (2025) 109576.

\bibitem{hautus1983strong}
M.~L. Hautus, Strong detectability and observers, Linear Algebra and its applications 50 (1983) 353--368.

\bibitem{shu2007detectability}
S.~Shu, F.~Lin, H.~Ying, Detectability of discrete event systems, IEEE Transactions on Automatic Control 52~(12) (2007) 2356--2359.

\bibitem{sanfelice2011convergence}
R.~G. Sanfelice, L.~Praly, Convergence of nonlinear observers on $\mathbb{R}^{n}$ with a riemannian metric (part i), IEEE Transactions on Automatic Control 57~(7) (2011) 1709--1722.

\bibitem{sanfelice2015convergence}
R.~G. Sanfelice, L.~Praly, Convergence of nonlinear observers on $\mathbb{R}^{n}$ with a riemannian metric (part ii), IEEE Transactions on Automatic Control 61~(10) (2015) 2848--2860.

\bibitem{sanfelice2023convergence}
R.~G. Sanfelice, L.~Praly, Convergence of nonlinear observers on $\mathbb{R}^{n}$ with a riemannian metric (part iii), IEEE Transactions on Automatic Control (2023).

\bibitem{amari2000methods}
S.-i. Amari, H.~Nagaoka, Methods of information geometry, Vol. 191, American Mathematical Soc., 2000.

\bibitem{alessandri2004design}
A.~Alessandri, Design of observers for lipschitz nonlinear systems using lmi, IFAC Proceedings Volumes 37~(13) (2004) 459--464.

\bibitem{shin2020convergence}
Y.~Shin, J.~Darbon, G.~E. Karniadakis, On the convergence of physics informed neural networks for linear second-order elliptic and parabolic type pdes, arXiv preprint arXiv:2004.01806 (2020).

\bibitem{kuri2014best}
A.~F. Kuri-Morales, The best neural network architecture, in: Nature-Inspired Computation and Machine Learning: 13th Mexican International Conference on Artificial Intelligence, MICAI 2014, Tuxtla Guti{\'e}rrez, Mexico, November 16-22, 2014. Proceedings, Part II 13, Springer, 2014, pp. 72--84.

\bibitem{oh2021optimal}
B.~K. Oh, J.~Kim, Optimal architecture of a convolutional neural network to estimate structural responses for safety evaluation of the structures, Measurement 177 (2021) 109313.

\bibitem{peralez2021deep}
J.~Peralez, M.~Nadri, Deep learning-based luenberger observer design for discrete-time nonlinear systems, in: 2021 60th IEEE Conference on Decision and Control (CDC), IEEE, 2021, pp. 4370--4375.

\bibitem{buisson2023towards}
M.~Buisson-Fenet, L.~Bahr, V.~Morgenthaler, F.~Di~Meglio, Towards gain tuning for numerical kkl observers, IFAC-PapersOnLine 56~(2) (2023) 4061--4067.

\bibitem{deng2023effect}
T.~Deng, Effect of the number of hidden layer neurons on the accuracy of the back propagation neural network, Highlights Sci. Eng. Technol 74 (2023) 462--468.

\bibitem{kingma2014adam}
D.~P. Kingma, J.~Ba, Adam: A method for stochastic optimization, arXiv preprint arXiv:1412.6980 (2014).

\bibitem{marino1993adaptive}
R.~Marino, S.~Peresada, P.~Valigi, Adaptive input-output linearizing control of induction motors, IEEE Transactions on Automatic control 38~(2) (1993) 208--221.

\bibitem{praly2006new}
L.~Praly, A.~Isidori, L.~Marconi, A new observer for an unknown harmonic oscillator, in: Proceedings of the 17th International Symposium on Mathematical Theory of Networks and Systems, Kyoto, Japan, 2006, pp. 24--28.

\bibitem{outbib1992stabilizability}
R.~Outbib, G.~Sallet, Stabilizability of the angular velocity of a rigid body revisited, Systems \& Control Letters 18~(2) (1992) 93--98.

\bibitem{boutayeb2000reduced}
M.~Boutayeb, M.~Darouach, A reduced-order observer for non-linear discrete-time systems, Systems \& control letters 39~(2) (2000) 141--151.

\bibitem{jammazi2021global}
C.~Jammazi, M.~Boutayeb, G.~Bouamaied, On the global polynomial stabilization and observation with optimal decay rate, Chaos, Solitons \& Fractals 153 (2021) 111447.

\end{thebibliography}

\end{document}